\documentclass{article}

\usepackage{fullpage}

\usepackage[utf8]{inputenc} % allow utf-8 input
\usepackage[T1]{fontenc}    % use 8-bit T1 fonts
\usepackage{hyperref}       % hyperlinks
\usepackage{url}            % simple URL typesetting
\usepackage{booktabs}       % professional-quality tables
\usepackage{amsfonts}       % blackboard math symbols
\usepackage{nicefrac}       % compact symbols for 1/2, etc.
\usepackage{microtype}      % microtypography
\usepackage{xcolor}         % colors
\usepackage{amsmath}
\usepackage{graphicx}
\usepackage{enumitem}
\newcommand\Real{{\mathbb R}}
\newcommand\vs{{\vspace*{-0.1cm}}}
\usepackage{multirow}
\usepackage{subcaption}

\newcommand\citep\cite

\title{GraphiT: Encoding Graph Structure in Transformers}

\author{
  Grégoire Mialon\footnote{Equal contribution.} \\ % todo put the joint first author mark
  Inria\footnote{Univ. Grenoble Alpes, Inria, CNRS, Grenoble INP, LJK, 38000 Grenoble, France.}\ \footnote{D.I., UMR 8548, École Normale Supérieure, Paris, France.} \\
  \texttt{gregoire.mialon@inria.fr}
  \and
  Dexiong Chen\footnotemark[1] \\
  Inria\footnotemark[2] \\
  \texttt{dexiong.chen@inria.fr}
  \and
  Margot Selosse\footnotemark[1] \\
  Inria\footnotemark[2] \\
  \texttt{margot.selosse@inria.fr}
  \and
  Julien Mairal \\
  Inria\footnotemark[2] \\
  \texttt{julien.mairal@inria.fr}
}

\begin{document}

\maketitle

\begin{abstract}
   % We introduce a new graph representation based on an optimal transport kernel for graphs. In particular, this kernel allows to adaptively pool the nodes of a graph. On the one hand, the kernel point of view offers an unsupervised and very expressive data representation, which is useful when limited labeled samples are available. On the other hand, our model can also be trained end-to-end when more data can be curated. Moreover, we thoroughly re-evaluate popular graph representations on several graph classification benchmarks, for which our method achieves competitive performance.
   We show that viewing graphs as sets of node features and incorporating structural and positional information into a transformer architecture is able to outperform representations learned with classical graph neural networks (GNNs). 
   Our model, GraphiT, encodes such information by (i) leveraging relative positional encoding strategies in self-attention scores based on positive definite kernels on graphs, and (ii) enumerating and encoding local sub-structures such as paths of short length. We thoroughly evaluate these two ideas on many classification and regression tasks, demonstrating the effectiveness of each of them independently, as well as their combination. In addition to performing well
   on standard benchmarks, our model also admits natural visualization mechanisms for interpreting graph motifs explaining the predictions, making it a potentially strong candidate for scientific applications where interpretation is important.\footnote{Code available at \url{https://github.com/inria-thoth/GraphiT}.}
 
   %Structural information may be encoded by enumerating local sub-structures such as paths of short length,
   %While bag of nodes approaches are already competitive with GNNs for simple datasets, 
   %
   %providing the transformer with positional information about the nodes in the graphs or a description of local structures, enables to outperform GNNs.
   %We thoroughly compare different ways of incorporating this information and propose a new method based on a family of kernels on graphs, in the spirit of relative positional encoding which has proven to be beneficial for classical transformers on sequence data. 
   %This put into question the capacity of GNNs to make use of the graphs structure, while paving the way for larger scale models for graph representation.
\end{abstract}

\section{Introduction} 

Graph-structured data are present in numerous scientific applications and are the subject of growing interest. Examples of such data are as varied as proteins in computational biology~\citep{senior2020improved}, which may be seen as a sequence of amino acids, but also as a graph representing their tertiary structure, molecules in chemoinformatics~\citep{duvenaud2015convolutional}, shapes in computer vision and computer graphics~\citep{verma2018feastnet}, electronic health records, or communities in social networks. Designing graph representations for machine learning is a particularly active area of research, even though not new~\citep{borgwardt2020graph}, with a strong effort currently focused on graph neural networks~\citep{bronstein17,Chen2020,kipf2017semisupervised,scarselli2008graph,Velickovic2018,xu2019powerful}. A major difficulty is to find graph representations that are computationally tractable, adaptable to a given task, and capable of distinguishing graphs with different topological structures and local characteristics.

%Graphs are a data modality from which valuable knowledge and applications could be built, should suitable models be designed to learn from these: physical systems, molecules or knowledge base can indeed be seen as graphs. Graph Neural Networks (GNNs), the current standard architecture for learning from graphs~\citep{kipf2017semisupervised, Velickovic2018, xu2019powerful, Chen2020}, have initially been derived as an extension of convolutions for neighboring nodes in the graph, thus trying to exploit its structure. In its most general form, a GNN layer aggregates features of neighboring nodes in a permutation equivariant way, with a variety of proposed aggregation schemes. %Does it scale? Does it make the optimal use of the graph's structure?

In this paper, we are interested in the transformer, which has become the standard architecture for natural language processing~\citep{Vaswani2017}, and is gaining ground in computer vision~\citep{dosovitskiy2021an} and computational biology~\citep{rives2019biological}. The ability of a transformer to aggregate information across long contexts for sequence data makes it an interesting challenger for GNNs that successively aggregate local information from neighbors in a multilayer fashion. In contrast, a single self-attention layer of the transformer can potentially allow all the nodes of an input graph to communicate. %The price to pay is that a naive implementation of the transformer for graphs requires seeing data points as sets of node features, while discarding the topological structure of the graph and thus losing crucial information.
The price to pay is that this core component is invariant to permutations of the input nodes, hence does not take the topological structure of the graph into account and looses crucial information. 
%, which is why it can be seen as a GNN operating on a fully-connected graph. For the same reason, using a bare Transformer on graphs may yield poor results, as the structure of the graph is not exploited.

For sequence data, this issue is addressed by encoding positional information of each token and giving it to the transformer architecture. For graphs, the problem is more challenging as there is no single way to encode node positions. For this reason, there have been few attempts to use vanilla transformers for graph representation. To the best of our knowledge, the closest work to ours seems to be the graph transformer architecture of Dwivedi and Bresson \cite{dwivedi2021generalization}, who 
propose an elegant positional encoding strategy based on the eigenvectors of the graph Laplacian~\citep{belkin2003}. However, they focus on applying attention to neighboring nodes only, as in GNNs, and their results suggest that letting all nodes communicate is not a competitive approach.

%Moreover, using these eigenvectors as position encoding raise different problems that will be detailed in the paper. 
%it is not clear whether eigenvectors of different graph Laplacian can be compared. Finally, the use of eigenvectors can be restrictive: in the public implementation, the length of the position encoding is limited by the size of the smallest graph of the data set.
%{\color{red} Julien: this is a very imprecise statement, and perhaps controversial. After all, their stuff works.}

Our paper provides another perspective and reaches a slightly different conclusion; we show that even though local communication is indeed often more effective, the transformer with global communication can also achieve good results. For that, we introduce a set of techniques to encode the local graph structure within our model, GraphiT (encoding \textbf{graph} structure \textbf{i}n \textbf{t}ransformers). More precisely, GraphiT relies on two ingredients that may be combined, or used independently.
First, we propose relative positional encoding strategies for weighting attention scores by using positive definite kernels, a viewpoint introduced for sequences in~\cite{tsai2019transformer}. This concept is particularly appealing for graphs since it allows to leverage the rich literature on kernels on graphs~\cite{kondor04,smola2003graphkernel}, which are powerful tools for encoding the similarity between nodes. The second idea consists in computing features encoding the local structure in the graph. To achieve this, we leverage the principle of graph convolutional kernel networks (GCKN) of~\cite{Chen2020}, which consists in enumerating and encoding small sub-structure (for instance, paths or subtree patterns), which may then be used as an input to the transformer model.

We demonstrate the effectiveness of our approach on several classification and regression benchmarks, showing that GraphiT with global or local attention layer can outperform GNNs in various tasks, and also show that basic visualization mechanisms allow us to automatically discover discriminative graph motifs, which is potentially useful for scientific applications where interpretation is important.

\section{Related work}\label{sec:related}

\paragraph{Graph kernels.} 
A classical way to represent graphs for machine learning tasks consists in defining a high-dimensional embedding for graphs, which may then be used to perform prediction with a linear models (\emph{e.g.}, support vector machines). Graph kernels typically provide such embeddings by counting the number of occurrences of local substructures that are present within a graph~\citep{borgwardt2020graph}. The goal is to choose substructures leading to expressive representations sufficiently discriminative, while allowing fast algorithms to compute the inner-products between these embeddings. For instance, walks have been used for such a purpose~\citep{gartner2003graph}, as well as shortest paths~\citep{borgwardt2005shortest}, subtrees~\citep{harchaoui2007image,mahe2009graph,shervashidze2011weisfeiler}, or graphlets~\citep{shervashidze2009efficient}. Our work uses short paths, but other substructures could be used in principle. Note that graph kernels used in the context of comparing graphs, is a line of work different from the kernels on graphs that we will introduce in Section~\ref{sec:preliminaries} for computing embeddings of nodes.

\vs
\paragraph{Graph neural networks.}
Originally introduced in~\cite{scarselli2008graph}, GNNs have been derived as an extension of convolutions for graph-structured data: they use a message passing paradigm in which vectors (messages) are exchanged (passed) between neighboring nodes whose representations are updated using neural networks. Many strategies have been proposed to aggregate features of neighboring nodes (see, \textit{e.g},~\cite{ bronstein17,duvenaud2015convolutional}). The graph attention network (GAT)~\citep{Velickovic2018} is the first model to use an attention mechanism for aggregating local information. Recently, hybrid approaches between graph neural networks and graph kernels were proposed in~\cite{Chen2020,du2019graph}. Diffusion processes on graphs that are related to the diffusion kernel we consider in our model were also used within GNNs in~\cite{klicpera2019diffusion}.

%\subsection{Transformers and Position encoding in NLP and vision}

\vs
\paragraph{Transformers for graph-structured data.} 

Prior to~\cite{dwivedi2021generalization}, there were some attempts to use transformers in the context of graph-structured data. The authors of~\cite{li2019} propose to apply attention to all nodes, yet without position encoding. In \cite{zhang2020graphbert}, a transformer architecture called Graph-BERT is fed with sampled subgraphs of a single graph in the context of node classification and graph clustering. They also propose to encode positions by aggregating different encoding schemes. However, these encoding schemes are either impossible to use in our settings as they require having sampled subgraphs of regular structures as input, or less competitive than Laplacian eigenvectors as observed in~\cite{dwivedi2021generalization}. The transformer model introduced in \cite{yun2019} needs to first transform an {heteregoneous} graph into a new graph structure via meta-paths, which does not directly operate on node features. To the best of our knowledge, our work is the first to demonstrate that vanilla transformers with appropriate node position encoding can compete with GNNs in graph prediction tasks. 

\section{Preliminaries about Kernels on Graphs}
\label{sec:preliminaries}

%\subsection{The Diffusion kernel}
%Our position encoding for nodes will be based on a family of kernel on graphs, which we now introduce.

\paragraph{Spectral graph analysis.}
 The Laplacian of a graph with $n$ nodes is defined as $L = D - A$, where~$D$ is a $n \times n$ diagonal matrix that carries the degrees of each node on the diagonal and $A$ is the adjacency matrix. Interestingly, $L$ is a positive semi-definite matrix such that for all vector $u$ in $\Real^n$, $u^\top L u = \sum_{i\sim j} (u[i]-u[j])^2$, which can be interpreted as the amount of ``oscillation'' of the vector~$u$, hence its ``smoothness', when seen as a function on the nodes of the graph.
 
 The Laplacian is often used via its eigenvalue decomposition $L= \sum_{i} \lambda_i u_i u_i^\top$, where the eigenvalue $\lambda_i = u_i^\top L u_i$ characterizes the amount of oscillation of the corresponding eigenvector~$u_i$. For this reason, this decomposition is traditionally viewed as the discrete equivalent to the sine/cosine Fourier basis in $\mathbb{R}^n$ and associated frequencies. Note that very often the normalized Laplacian $I - D^{-\frac{1}{2}} A D^{-\frac{1}{2}}$ is used instead of $L$, which does not change the above interpretation.
 
Interestingly, it is then possible to define a whole family of positive definite kernels on the graph~\citep{smola2003graphkernel} by applying a regularization function $r$ to the spectrum of $L$. We therefore get a rich class of kernels
\begin{equation}
    K_r = \sum_{i=1}^m {r(\lambda_i)} u_i u_i^\top,
\end{equation}
associated with the following norm $\|f\|_r^2 = \sum_{i=1}^m {({f}_i^\top u_i)^2}/{r(\lambda_i)}$ from a reproducing kernel Hilbert space (RKHS),
where $r : \mathbb{R} \mapsto \mathbb{R}_*^+ $ is a non-increasing function such that smoother functions on the graph would have smaller norms in the RKHS. We now introduce two examples. 

%\paragraph{Laplacian kernel.} A first example of this family is obtained by taking $r(\lambda_i) = \frac{1}{\lambda_i + \epsilon}$, with $\epsilon \geq 0$. Then, we get the regularized Laplacian Kernel:
%\begin{equation}
%    K = (L + \epsilon I)^{-1}.
%\end{equation}

\vs
\paragraph{Diffusion Kernel~\cite{kondor04}.} 
It corresponds to the case $r(\lambda_i) = e^{-\beta \lambda_i}$, which gives:
\begin{equation}
    K_{D} = \sum_{i=1}^m e^{- \beta \lambda_i} u_i u_i^\top = e^{- \beta L} = \lim_{p \to +\infty} \left(  I - \frac{\beta}{p}L\right)^p.\label{eq:diffusion}
\end{equation}
The diffusion kernel can be seen as a discrete equivalent of the Gaussian kernel, a solution of the heat equation in the continuous setting, hence its name. Intuitively, the diffusion kernel between two nodes can be interpreted as the quantity of some substance that would accumulate at the first node after a given amount of time (controlled by $\beta$) if we injected the substance at the other node and let it diffuse through the graph. It is related to the random walk kernel that will be presented below.

%By definition of matrix exponentiation:
%\begin{equation}
%    K_{D} = \sum_{t=0}^{+\infty} (- \beta)^t \frac{L^t}{t!}.
%\end{equation}
%Hence, $K_D$ spans all the graph, whereas the $p$-step Random Walk kernel has a span of p neighbors.
%Members of this family can be seen as low-pass filters: the higher $\beta$, the more high frequencies (\textit{i.e}, high $\lambda_i$) are blocked.

\vs
\paragraph{p-step random walk kernel.} 
By taking $r(\lambda_i) = (1 - \gamma \lambda_i)^{p}$ , we obtain a kernel that admits an interpretation in terms of p steps of a particular random walk on the graph:
\begin{equation}
    K_{pRW} = (I  - \gamma L)^{p}.
\end{equation}
It is related to the diffusion kernel by choosing $\gamma=\beta/p$ and taking the limit with $p \to +\infty$, according to~(\ref{eq:diffusion}). Unlike the diffusion kernel which yields a dense matrix, the random walk kernel is sparse and has limited connectivity, meaning that two nodes are connected in $K_{pRW}$ only if there exists a path of length~$p$ between them.
As these kernels on graphs reflect the structural similarity of the nodes independently of their features,
%(note how substructures appear on Figure~\ref{fig:ex_diff}),
it is natural to use the resulting Gram matrix to encode such a structure within the transformer model, as detailed next.

\section{Encoding Graph Structure in Transformers}

In this section, we detail the architecture and structure encoding strategies behind GraphiT. In particular, we build upon the findings of~\cite{tsai2019transformer} for our architectural choices and propose new strategies for encoding structural information in the transformer architecture.

\subsection{Transformer Architectures for Graphs}

We process graphs with a vanilla transformer encoder architecture~\citep{Vaswani2017} for solving classification and regression tasks, by seeing graphs as sets of node features. 
We first present the transformer without encoding the graph structure, before introducing mechanisms for encoding nodes positions in Section~\ref{subsec:position}, and then topological structures in Section~\ref{subsec:structure}.
Specifically, a transformer is composed of a sequence of layers, which process an input set of $d_{\text{in}}$ features represented by a matrix ${X}$ in $\Real^{n \times d_{\text{in}}}$, and compute another set in $\Real^{n \times d_{\text{out}}}$. A critical component is the attention mechanism:
\begin{equation}
    \text{Attention}(Q, V) = \text{softmax}\left( \frac{QQ^\top}{\sqrt{d_{out}}} \right) V \in \Real^{n \times d_{\text{out}}},
    \label{eq:attention}
\end{equation}
with $Q^\top = W_Q {X}^\top$ is called the query matrix, $V^\top = W_V {X}^\top$ the value matrix, and $W_Q, W_V$ in $\Real^{d_{\text{out}} \times d_{\text{in}}}$ are projection matrices that are learned. Note that following the recommendation of~\cite{tsai2019transformer}, we use the same matrix for keys and queries. This allows us to define a symmetric and positive definite kernel on pairs of nodes and thus to combine with other kernels on graph. This also reduces the number of parameters in our models without hurting the performance in practice. During the forward pass, the feature map $X$ is updated in a residual fashion (with either $d_{\text{in}}=d_{\text{out}}$ or with an additional projection matrix when the dimensions do not match, omitted here for simplicity) as follows:
\begin{equation*}
    {X} = {X} + \text{Attention}(Q, V).
    % \label{eq:pos_forward_ori}
\end{equation*}

Note that transformers without position encoding and GNNs are tightly connected: a GNN can be seen as a transformer where aggregation is performed on neighboring nodes only, and a transformer as a GNN processing a fully-connected graph. However, even in this case, differences would remain between common GNNs and Transformers, as the latter use for example LayerNorm and skip-connections. In our paper, we will either adopt the local aggregation strategy, or let all nodes communicate in order to test this inductive bias in the context of graphs, where capturing long range interactions between nodes may be useful for some tasks. 

\subsection{Encoding Node Positions}\label{subsec:position}

The output of the transformer is invariant to permutations in the input data. It is therefore crucial to provide the transformer with information about the data structure. In this section, we revisit previously proposed strategies for sequences and graphs and devise new methods for positional encoding. Note that a natural baseline is simply to adopt a local aggregation strategy similar to GAT~\cite{Velickovic2018} or~\cite{dwivedi2021generalization}.
% and draw links with existing strategies in Transformers for Graphs and Natural Language Processing.

\subsubsection{Existing Strategies for Sequences and Graphs}

\paragraph{Absolute and relative positional encoding in transformers for sequences.} % Explain the difference and that relative PE found to work better in NLP.
In NLP, positional information was initially encoded by adding a vector based on sinusoidal functions to each of the input token features~\cite{Vaswani2017}.
This approach was coined as absolute position encoding and proved to be also useful for other data modalities~\citep{dosovitskiy2021an,dwivedi2021generalization}. In relative position encoding, which was proposed later~\citep{parikh-etal-2016-decomposable, shaw2018selfattention}, positional information is added to the attention logits and in the values. This information only depends on the relative distance between the two considered elements in the input sequence. 
%In general, position encoding is either hand-crafted~\cite{Vaswani2017} or learned~\cite{Vaswani2017,dosovitskiy2021an,sha} while absolute position encoding is always learned~\citep{shaw2018selfattention}.

%In vision When feeding a Transformer with image patches, \cite{dosovitskiy2021an} observe that 1D \textit{learned} position encoding added to the input patches is sufficient compared to other position encodings. Their interpretation is that the Transformer operates at a patch level with a relatively small number of patches, hence with small spatial dimensions for which learning a position encoding is easy. 

%\paragraph{Point clouds.} In the context of vector attention, \cite{zhao2020point} use a trainable MLP with two linear layers and one ReLU nonlinearity to embed the difference between the $3$D coordinates of two points. This is therefore an instance of relative position encoding. The output is then added to the regular attention before applying the softmax, and to the value. The position encoding is trained end-to-end with the rest of the network. In their experiments, relative position encoding performs better than absolute position encoding.

\paragraph{Absolute positional encoding for graphs.}

Whereas positional encodings can be hand-crafted or learned relatively easily for sequences and images, which respectively admit a chain or grid structure, this task becomes non-trivial for graphs, whose structure may vary a lot inside a data set, besides the fact that the concept of node position is ill-defined. To address these issues, an absolute position encoding scheme was recently proposed in \cite{dwivedi2021generalization}, by leveraging the graph Laplacian (LapPE). More precisely, each node of each graph is assigned a vector containing the first $k$ coordinates of the node in the eigenbasis of the graph normalized Laplacian sorted in ascending order of the eigenvalues. Since the first eigenvector associated to the eigenvalue 0 is constant, the first coordinate is omitted.
%The graph Laplacian is defined as $L = D - A$, where $D$ is the degree matrix and $A$ the adjacency matrix of the graph. If we diagonalize the normalized Laplacian $\Delta$,
%\begin{equation}
 %   \Delta = I - D^{-\frac{1}{2}} A D^{-\frac{1}{2}} = U \Lambda U^\top,
%\end{equation}
%we can use the eigenvectors (columns of $U$) corresponding to the $k$ smallest non-trivial eigenvalues in $\Lambda$ as a coordinate system for the nodes position (for a graph with a single connected component, $0$ is an eigenvalue of multiplicity one of $L$).
%The eigenvectors can be seen as the discrete equivalent of the Fourier basis in $\mathbf{R}^n$.

As detailed in Section~\ref{sec:preliminaries}, these eigenvectors oscillate more and more and the corresponding coordinates are often interpreted as Fourier coefficients representing frequency components in increasing order. Note that eigenvectors of the Laplacian computed on different graphs could not be compared to each other in principle, and are also only defined up to a $\pm 1$ factor.
While this raises a conceptual issue for using them in an absolute positional encoding scheme, it is shown in~\cite{dwivedi2021generalization}---and confirmed in our experiments---that the issue is mitigated by the Fourier interpretation, and that
the coordinates used in LapPE are effective in practice for discriminating between nodes in the same way as the position encoding proposed in~\cite{Vaswani2017} for sequences. 
Yet, because the eigenvectors are defined up to a $\pm$1 factor, the sign of the encodings needs to be randomly flipped during the training of the network. %Both~\cite{dwivedi2021generalization} and our experiments suggest that LapPE improves the performance of a vanilla Transformer attending on all the nodes of an input graph. 
%However, it is not clear whether the encodings can indeed be transfered from graph to graph. 
In the next section, we introduce a novel strategy in the spirit of relative positional encoding that does not suffer from the previous conceptual issue, and can also be combined with LapPE if needed.

%Second, in their implementation, the dimension of the positional encoding is limited by the size of the smallest graph in the data set, which can be restrictive (the dimension of the Laplacian is equal to the number of nodes in the graph). For example, the smallest graph in PTC has size $2$, and $3$ in PROTEINS.  Third, the sign of the eigenvectors is defined up to a sign. The sign is therefore flipped during the training. But it seems to prevent convergence of the solver in the kernel setting. %TODO: check arguments of the FAIR paper against Laplacian eigenvectors.

%\paragraph{Weisfeiler-Lehman PE.} \cite{dwivedi2021generalization} mention another possible position encoding introduced in~\cite{zhang2020graphbert}. They rely on the Weisfeiler-Lehman algorithm to get a code for each node of the graph ($WL(n_j)$). Then, similarly to~\cite{Vaswani2017}, they compute the following vector:
%\begin{equation}
 %   e(n_j) = PositionEmbed(WL(n_j)) \\
 %   = \Big[sin\Big(\frac{WL(n_j)}{10000^{\frac{2l}{d_h}}}\Big), cos\Big(\frac{WL(n_j)}{10000^{\frac{2l}{d_h}}}\Big)\Big]_{l = 0}^{\frac{d_h}{2}}
%\end{equation}

%\paragraph{Problems with WL PE.} It is outperformed by Laplacian PE in the experiments of~\cite{dwivedi2021generalization}.

\subsubsection{Relative Position Encoding Strategies by Using Kernels on Graphs}

%\paragraph{GraphWave as absolute positional encoding.} Our first attempt to better integrate positional information consists in using GraphWave as an absolute positional encoding: we add to each node feature vector another feature vector containing the node's positional information. This is the approach adopted by~\cite{Vaswani2017} and \cite{dwivedi2021generalization}, although with different positional information vectors. The GraphWave embeddings can be pre-computed and the dimension chosen arbitrarily, thus making its use practical. Finally, GraphWave is theoretically motivated. In particular, \cite{Donnat_2018} show that the spectral graph wavelet of a given node is tightly linked to the topological properties of its local neighborhood. According to the authors, GraphWave transfers well between graphs.
%outperforms all existing methods for structural node embedding. The task consists in predicting nodes structural role (manually labeled) in a synthetic graph data set using the structural embedding.

%Need for relative position encoding because the problem of absolute position encoding is non-trivial, and/or such PE should be difficult to learn for graphs, as opposed to regular structure such as sequences in NLP and grid in Computer Vision.

\paragraph{Modulating the transformer attention.} 
To avoid the issue of transferability of the absolute positional encoding between graphs, we use information on the nodes structural similarity to bias the attention scores. More precisely, and in the fashion of~\cite{Mialon2021,tsai2019transformer} for sequences, we modulate the attention kernel using the Gram matrix of some kernel on graphs described in Section~\ref{sec:preliminaries} as follows:
\begin{equation}
    \text{PosAttention}(Q, V, K_r) = \text{normalize}\left(\text{exp}\left( \frac{QQ^\top}{\sqrt{d_{\text{out}}}}\right) \odot K_r \right) V \in \Real^{n \times d_{\text{out}}},
    \label{eq:pos_attention}
\end{equation}
with the same $Q$ and $V$ matrices, and $K_r$ a kernel on the graph. ``$\text{exp}$'' denotes the elementwise exponential of the matrix entries and $\text{normalize}$ means $\ell_1$-normalization on rows such that normalization(exp(u))=softmax(u). The reason for multiplying the exponential of the attention logits before $\ell_1$-normalizing the rows is that it corresponds to a classical kernel smoothing. Indeed, if we consider the PosAttention output for a node $i$:
\begin{equation*}
    \text{PosAttention}(Q, V, K_r)_i = \sum_{j=1}^n \frac{\text{exp}\left(\nicefrac{Q_i Q_j^\top}{\sqrt{d_{out}}} \right) \times K_r(i, j)}{\sum_{j'=1}^n \text{exp}\left(\nicefrac{Q_i Q_{j'}^\top}{\sqrt{d_{out}}} \right) \times K_r(i, j')} V_j \in \Real^{d_{\text{out}}},
    \label{eq:pos_attention_i}
\end{equation*}
we obtain a typical smoothing, \textit{i.e}, a linear combination of features with weights determined by a non-negative kernel, here $k(i, j) := \text{exp} (\nicefrac{Q_i Q_j^\top}{\sqrt{d_{out}}}) \times K_r(i, j)$, and summing to 1. In fact, $k$ can be considered as a new kernel between nodes $i$ and $j$ made by multiplying a kernel based on positions ($K_r$) and a kernel based on content (via $Q$). As observed in~\cite{tsai2019transformer} for sequences, modulating the attention logits with a kernel on positions is related to relative positional encoding~\citep{shaw2018selfattention}, where we bias the attention matrix with a term depending only on the relative difference in positions between the two elements in the input set. Moreover, during the forward pass, the feature map ${X}$ is updated as follows:
\begin{equation}
    {X} = {X} + D^{-\frac{1}{2}} \text{PosAttention}(Q, V, K_r),
    \label{eq:pos_forward}
\end{equation}
where $D$ is the matrix of node degrees. We found such a normalization with $D^{-1/2}$ to be beneficial in our experiments since it reduces the overwhelming influence of highly connected graph components.
Note that, as opposed to  absolute position encoding, we do not add positional information to the values and the model does not rely on the transferability of eigenvectors between different Laplacians.

%, as~\cite{tsai2019transformer} observe that this does not significantly change the performance of the subsequent model.
 
 \paragraph{Choice of kernels and parameters.} Interestingly, the choice of the kernel enables to encode a priori knowledge directly within the model architecture, while the parameter has an influence on the attention span. For example, in the diffusion kernel, $\beta$ can be seen as the duration of the diffusion process. The smaller it is, the more focused the attention on the close neighbors. Conversely, a large~$\beta$ corresponds to an homogeneous attention span. As another example, it is clear that the choice of $p$ in the random walk kernel corresponds to visiting at best $p$-degree neighbors. In our experiments, the best kernel may vary across datasets, suggesting that long-range global interactions are of different importance depending on the task. For example, on the dataset PROTEINS (see Section~\ref{sec:experiments}), the vanilla transformer without using any structural information performs very well.

%\begin{itemize}

%\item This kernel can be multiplied pointwise with the attention scores matrix $A$:
%\begin{equation}
%    A = A * e^{- \beta L}.
%\end{equation}
%his can be interpreted as multiplying two kernels.

%\item In relative positional encoding, positional information is added to the attention scores. More precisely, in~\cite{shaw2018selfattention}, the edge between input elements $x_i$ and $x_j$ is represented by vectors $a_{ij}^V$, $a_{ij}^K \in \mathbb{R}^{d_a}$, and the information propagated in two places in the network: 
%\begin{equation}
%    z_i = \sum_{j=1}^n \alpha_{ij}(x_jW^V + a_{ij}^V),
%\end{equation}
%and
%\begin{equation}
%    e_{ij} = \frac{x_i W^Q(x_j W^K + a_{ij}^K)^\top}{\sqrt{d_z}},
%\end{equation}
%with $\alpha_{ij} = \frac{exp e_{ij}}{\sum_{k=1}^n exp e_{ik}}$, $z_i$ being the feature map of the element $i$. Could we replace $a_{ij}$ by $K_{ij}$? In the Transformer, the $a_{ij}$'s are learned.

%\item Finally, given a feature map $Z \in \mathbb{R}^{n \times d}$ (with $n$ the number of elements in the input sequence and $d$ the hidden dimension), we could do the following matrix multiplication:
%\begin{equation}
%    Z = K Z,
%\end{equation}
%which can be interpreted as using a low-pass filter on each dimension of the feature map, seen as a function on the graph.
%\end{itemize}

\subsection{Encoding Topological Structures} \label{subsec:structure}

%TODO: clarify difference between node position encoding and structure encoding.
Position encoding aims at adding positional only information to the feature vector of an input node or to the attentions scores. Substructures is a different type of information, carrying local positional information and content, which has been heavily used within graph kernels, see Section~\ref{sec:related}. 
%It has been shown in computer vision~\citep{dosovitskiy2021an} that feeding a transformer with pre-computed substructures can be a very effective strategy. Indeed, the Vision Transformer (ViT) is fed with linear projections of image patches, an inductive bias reminiscent of convolutions. In this case, the inputs bear both content and structural information. 
In the context of graphs neural networks, this idea was exploited in the graph convolutional kernel network model (GCKN) of~\cite{Chen2020}, which is a hybrid approach between GNNs and graph kernels based on substructure enumeration (\textit{e.g.}, paths). Among different strategies we experimented, enriching nodes features with the output of a GCKN layer turned out to be a very effective strategy.

%is based on path enumerations, encoding of path information, and local aggregation. More precisely, 
%we propose to feed our model with pre-computed paths. More precisely, and starting from a node, we concatenate the node features of along the path. Once this pre-processing operation is done, we obtain new features on which we train our models. Intuitively, the paths should bring useful structural information to the transformer. As in the ViT architecture, node positional encoding can be used complementarily. As our first attempt to feed our models with concatenated features was inconclusive, we employed a more refined method, namely a Graph Convolutionnal Kernel Network layer~\citep{Chen2020}, to encode paths.

\paragraph{Graph convolutional kernel networks (GCKN).}
 GCKNs~\cite{Chen2020} is a multi-layer model that produces a sequence of graph feature maps akin to a GNN. The main difference is that each layer enumerates local sub-structures at each node (here, paths of length~$k$), encodes them using a kernel embedding, and aggregates the resulting sub-structure representations. This results in a feature map that carries more information about the graph structure than traditional neighborhood aggregation based GNNs, which is appealing for transformers since the vanilla version is blind to the graph structure.
 
 Formally, let us consider a graph $G$ with $n$ nodes, and let us denote by $\mathcal{P}_k(u)$ the set of paths shorter than or equal to $k$ that start with node $u$. With an abuse of notation, $p$ in $\mathcal{P}_k(u)$ will denote the concatenation of all node features encountered along the path. Then, a layer of GCKN defines a feature map 
 $X$ in $\Real^{n \times d}$ such that 
\begin{equation*}
    X(u)=\sum_{p \in \mathcal{P}_k(u)}  \psi(p),
\end{equation*}
where $X(u)$ is the column of $X$ corresponding to node $u$ and $\psi$ is a $d$-dimensional
embedding of the path features~$p$. More precisely, the path features in $\cite{Chen2020}$ are embedded to a RKHS by using a Gaussian kernel, and a finite-dimensional approximation is obtained by using the Nystr\"om method~\cite{williams2001using}. The embedding is parametrized and can be learned without or with supervision (see~\cite{Chen2020} for details). Moreover, path features of varying lengths up to a maximal length can be used with GCKN. In this work, we evaluate the strategy of encoding a node as the concatenation of its original features and those produced by one GCKN layer. This strategy has proven to be very successful in practice.

% A simple way to use the paths as input to the Transformer is to fix the length of the paths and train the transformer by considering the length as a hyper-parameter. However, using only one type of path is suboptimal and may even leave out essential structural information of the graphs. Another option is to consider all paths that are shorter than or equal to a fixed length. Then, a graph is seen as a set of paths of different lengths,  whose representations do not have the same dimension, which is an issue to train the transformer. The graph convolutional kernel network (GCKN) framework tackles this problem 

%\subsection{Computational Efficiency and Implementation}
%A computational advantage of using a Transformer instead of a GNN for processing graph structured data is that the complexity does not depend on the density of the graph, given that the position encoding is pre-computed. TODO: Discuss implementations. The datasets we consider do not contain large graphs (more than $1000$ nodes). As the cost of self-attention is quadratic in the size of the input sequence, considering large sequence as the input can be considered as problematic at first sight. However, a recent line of work coined as efficient Transformers largely alleviated these issues for large inputs both in terms of memory and computational cost~\citep{dai2019transformerxl,Beltagy2020Longformer,kitaev2020reformer}. We refer the interested reader to the following survey~\citep{tay2020efficient}.

\section{Experiments}
\label{sec:experiments}

In this section, we evaluate instances of GraphiT as well as popular GNNs and position encoding baselines on various graph classification and regression tasks. We want to answer several questions:
\begin{itemize}[leftmargin=0.7cm,itemsep=0pt,parsep=0pt,topsep=0pt]
    \item[\textbf{Q1}:] Can vanilla transformers, when equipped with appropriate position encoding and/or structural information, outperform GNNs in graph classification and regression tasks?
    \item [\textbf{Q2}:] Is kernel-based relative positional encoding more effective than the absolute position encoding provided by the eigenvectors of the Laplacian (LapPE)?
    %\item [\textbf{Q3}:] Should we use global or local aggregation strategies within graph transformers?
    \item [\textbf{Q3}:] What is the most effective way to encode graph structure information within transformers? 
\end{itemize}
We also discuss our results and conduct ablation studies. Finally, we demonstrate the ability of attention scores to highlight meaningful graph features when using kernel-based positional encoding.

\subsection{Methodology}

\paragraph{Benchmark and baselines.} %What do we do and why?
We benchmark our methods on various graph classification datasets with discrete node labels (MUTAG, PROTEINS, PTC, NCI1) and one regression dataset with discrete node labels (ZINC). These datasets can be obtained \textit{e.g} via the Pytorch Geometric toolbox~\citep{Fey2019}. %Which baselines? Why do we pick them?
We compare our models to the following GNN models: Molecular Fingerprint (MF)~\citep{duvenaud2015convolutional}, Graph Convolutional Networks (GCN)~\citep{kipf2017semisupervised}, Graph Attention Networks (GAT)~\citep{Velickovic2018}, Graph Isomorphism Networks (GIN)~\citep{xu2019powerful} and finally Graph Convolutional Kernel Networks (GCKN)~\citep{Chen2020}. In particular, GAT is an important baseline as it uses attention to aggregate neighboring node information. We compare GraphiT to the transformer architecture proposed in~\cite{dwivedi2021generalization} and also use their Laplacian absolute position encoding as a baseline for evaluating our graph kernel relative position encoding. 
%We evaluate our node position encoding and structural encoding methods separately and jointly. 
All models are implemented in Pytorch and our code is available in the supplementary material.

%Attention mechanisms in graphs have been proposed to adaptively aggregate neighbor's features~\citep{Velickovic2018}. However, attention is only applied to neighbors. Considering all nodes would probably be simpler in terms of implementation and scalability of the model, but requires to integrate positional information on the nodes in a graph. Learning complicated mechanisms may not be feasible when few labeled data is available (this is linked to the evaluation problem: many popular models are overfitting)? Going further, couldn't we simply consider a graph as a bag of node features, and integrating positional information?

\paragraph{Reporting scores.} %Many popular models for graph representations are selected and evaluated on the same folds. Many papers therefore report their score on validation set, which is not enough to conclude as of which method generalizes better to unseen data. %For example, the respective performance of the models introduced in~\cite{Chen2020} change when model selection and model evaluation are done separately. It seems that many popular models are able to overfit the validation set in many cases. A paper from last year tried to properly evaluate GNNs~\citep{Errica2020A}.
For all datasets except ZINC, we samples ten times random train/val/test splits, of size $80/10/10$, respectively. 
For each split, we evaluate all methods by (i) training several models on the train fold with various hyperparameters; (ii) performing model selection on val, by averaging the validation accuracy of a model on its last 50 epochs; (iii) retraining the selected model on train+val; (iv) estimate the test score by averaging the test accuracy over the last 50 epochs. 
%{nested} 10-fold cross-validation. More precisely, for each training fold, we perform model selection by dividing the fold into an inner train fold and an inner validation fold. Then, we select the set of hyper-parameters which performed the best on the validation, retrain the model on the inner train and validation fold, and report the accuracy on the corresponding test fold. 
The results reported in our tables are then averaged over the ten splits.
This procedure is a compromise between a double-nested cross validation procedure that would be too computationally expensive and reporting 10-fold cross validation scores that would overestimate the test accuracy.
%splitting the data set into a train, validation and test, which can output biased scores, and performing an exact nested cross-validation, which is unrealistic from a computational point of view for most datasets and methods in this work. The models are selected and the test score reported by averaging the accuracies over the last $50$ epochs of the training. Finally, the scores reported in the results section are the average test scores on the $10$ folds. 
For ZINC, we use the same train/val/test splits as in~\cite{dwivedi2021generalization}, train GraphiT with 10 layers, 8 heads and 64 hidden dimensions as in~\cite{dwivedi2021generalization}, and report the average test mean absolute error on 4 independent runs.

\paragraph{Optimization procedure and hyperparameter search.} 
Our models are trained with the Adam optimizer by decreasing the learning rate by a factor of 2 each 50 epochs. For classification tasks, we train about the same number (81) of models with different hyperparameters for each GNN and transformer method, thus spending a similar engineering effort on each method.
For GNN models, we select the best type of global pooling, number of layers and hidden dimensions from three different values. Regarding transformers, we select the best number of heads instead of global pooling type for three different values. For all considered models, we also select the best learning rate and weight decay from three different values and the number of epochs is fixed to 300 to guarantee the convergence. For the ZINC dataset, we found that a standard warmup strategy suggested for transformers in \cite{Vaswani2017} leads to more stable convergence for larger models. The rest of the hyperparameters remains the same as used in~\cite{dwivedi2021generalization}.
More details and precise grids for hyperparameter search can be found in Appendix~\ref{sec:add_exp}.

\subsection{Results and Discussion}

\begin{table}[t]
    \small
    \centering
    \caption{Average mean classification accuracy/mean absolute error.
    }
    \resizebox{\textwidth}{!}{
    \begin{tabular}{l|c|c|c|c|c} \toprule
    {Method / Dataset} & {MUTAG} & {PROTEINS} & {PTC} & {NCI1} & {ZINC (no edge feat.)} \\
    \midrule
    {Size} & {188} & {1113} & {344} & {4110} & {12k} \\
    {Classes} & {2} & {2} & {2} & {2} & {Reg.} \\
    {Max. number of nodes} & {28} & {620} & {109} & {111} & {37} \\
    \midrule
    {MF~\citep{duvenaud2015convolutional}} & {81.5$\pm$11.0} & {71.9$\pm$5.2} & {57.3$\pm$6.9} & {80.6$\pm$2.5} & {0.387$\pm$0.019} \\
    {GCN~\citep{kipf2017semisupervised}} & {78.9$\pm$10.1} & {75.8$\pm$5.5} & {54.0$\pm$6.3} & {75.9$\pm$1.6}  & {0.367$\pm$0.011} \\
    {GAT~\citep{Velickovic2018}} & {80.3$\pm$8.5} & {74.8$\pm$4.1} & {55.0$\pm$6.0} & {76.8$\pm$2.1}  & {0.384$\pm$0.007} \\
    {GIN~\citep{xu2019powerful}} & {82.6$\pm$6.2} & {73.1$\pm$4.6} & {55.0$\pm$8.7} & {\textbf{81.7$\pm$1.7}}  & {0.387$\pm$0.015} \\
    %{GCKN unsup. (subtree, no aggregation)} & {85.0 $\pm$ 10.2} & {72.8 $\pm$ 5.6} & {61.2 $\pm$ 8.4} & {TODO}  & {TODO} \\
    {GCKN-subtree~\citep{Chen2020}} & {87.8$\pm$9.4} & {72.0$\pm$3.7} & {\textbf{62.1$\pm$6.4}} & {79.6$\pm$1.8}  & {0.474$\pm$0.001} \\ % sup. (sub + agg, $k=4$, $\sigma=0.5$, sum)
     %{Path kernel~\citep{Chen2020}} & {83.3 $\pm$ 10.8} & {74.2 $\pm$ 4.5} & {57.9 $\pm$ 8.2} & {-}  & {-} \\
    \midrule
    %{MF + OTKE sup.} & {84.4 $\pm$ 10.2} & {70.1 $\pm$ 6.9} & {57.4 $\pm$ 6.6} & {79.0 $\pm$ 1.4}  & {TODO} \\
    %{GCN + OTKE sup.} & {74.2 $\pm$ 15.6} & {70.9 $\pm$ 6.3} & {56.9 $\pm$ 7.9} & {74.2 $\pm$ 2.2}  & {TODO} \\
    %{GAT + OTKE sup.} & {71.6 $\pm$ 13.4} & {71.4 $\pm$ 5.1} & {56.2 $\pm$ 7.8} & {75.8 $\pm$ 2.6}  & {TODO} \\
    %{GIN + OTKE sup.} & {79.9 $\pm$ 12.6} & {72.9 $\pm$ 5.2} & {57.5 $\pm$ 4.9} & {81.2 $\pm$ 1.5}  & {TODO} \\
    %{GCKN + OTKE unsup. (one layer, aggregation)} & {82.8 $\pm$ 8.5} & {72.3 $\pm$ 5.1} & {63.8 $\pm$ 7.2} & {TODO}  & {TODO} \\
    %{GCKN + OTKE sup.} & {TODO} & {TODO} & {TODO} & {TODO}  & {TODO} \\
    %\midrule
    %{Match kernel (gaussian)} & {82.8 $\pm$ 8.5} & {73.2 $\pm$ 3.6} & {56.5 $\pm$ 10.2} & {-}  & {-} \\
    %{Match kernel (gaussian) + Laplacian PE} & {NO CONV.} & {NO CONV.} & {NO CONV.} & {-}  & {-} \\
    %{Match kernel (gaussian) + GraphWave PE} & {86.7 $\pm$ 7.5} & {74.1 $\pm$ 3.9} & {59.1 $\pm$ 10.7} & {-}  & {-} \\
    %{OTKE + Diffusion PE unsup.} & {TODO} & {TODO} & {TODO} & {TODO} \\
    %{OTKE + Diffusion PE sup.} & {TODO} & {TODO} & {TODO} & {TODO} \\
    \citep{dwivedi2021generalization} & 79.3$\pm$11.6 & 65.8$\pm$3.1 & 58.4$\pm$8.2 & 78.9$\pm$1.1 & 0.359$\pm$0.014 \\
    \citep{dwivedi2021generalization} + LapPE & 83.9$\pm$6.5 & 70.1$\pm$3.2 & 57.7$\pm$3.1 & 80.0$\pm$1.9 & 0.323$\pm$0.013 \\
    \midrule
    {Transformers (T)} & 82.2$\pm$6.3 & 75.6$\pm$4.9 & 58.1$\pm$10.5 & 70.0$\pm$4.5  & 0.696$\pm$0.007 \\
    {T + LapPE} & 85.8$\pm$5.9 & 74.6$\pm$2.7 & 55.6$\pm$5.0 & 74.6$\pm$1.9 & 0.507$\pm$0.003 \\
    %{T + GraphWave PE} & 83.0$\pm$11.5 & 73.5$\pm$5.8 & 56.3$\pm$8.1 & 75.1$\pm$4.3  & {TODO} \\
    %{T + GraphWave PE (dim=dim-hidden)} & {84.1 $\pm$ 8.3} & {73.4 $\pm$ 5.0} & {58.3 $\pm$ 8.2} & {74.6 $\pm$ 3.8}  & {TODO} \\
    {T + Adj PE} & 87.2$\pm$9.8 & 72.4$\pm$4.9 & 59.9$\pm$5.9 & 79.7$\pm$2.0 & 0.243$\pm$0.005\\
    {T + 2-step RW kernel} & 85.3$\pm$6.9 & 72.8$\pm$4.5 & 62.0$\pm$9.4 & 78.0$\pm$1.5 & 0.243$\pm$0.010\\
    {T + 3-step RW kernel} & 83.3$\pm$6.3 & \textbf{76.2$\pm$4.4} & 61.0$\pm$6.2 & 77.6$\pm$3.6 & 0.244$\pm$0.011\\
    {T + Diffusion kernel} & 82.7$\pm$7.6 & 74.6$\pm$4.2 & 59.1$\pm$7.4 & 78.9$\pm$1.6  & {0.255$\pm$0.010} \\
    %{T + Diffusion kernel, $\beta = 0.1$} & {84.6 $\pm$ 9.0} & {74.2 $\pm$ 3.9} & {57.0 $\pm$ 8.7} & {70.9 $\pm$ 2.6}  & {TODO} \\
    %{T + Diffusion kernel, $\beta = 10$} & {85.0 $\pm$ 8.9} & {72.1 $\pm$ 4.4} & {56.6 $\pm$ 6.3} & {75.7 $\pm$ 2.4}  & {TODO} \\
    %{Path Transformers (PT)} & 80.1$\pm$10.4 & 73.1$\pm$4.3 & 58.3$\pm$11.9 & 71.4$\pm$2.0  & 0.669$\pm$0.003 \\
    {T + GCKN } & {84.4$\pm$7.8} & {69.5$\pm$3.8} & {61.5$\pm$5.8} & {78.1$\pm$5.1}  & 0.274$\pm$0.011 \\
    {T + GCKN + 2-step RW kernel} & {90.4$\pm$5.8} & {72.5$\pm$4.6} & {58.4$\pm$7.6} & {81.0$\pm$1.8} & {0.213$\pm$0.016}\\
    {T + GCKN + Adj PE} & {\textbf{90.5$\pm$7.0}} & {71.1$\pm$6.9} & {57.9$\pm$4.2} & {81.4$\pm$2.2} & {\textbf{0.211$\pm$0.010}}\\
    \bottomrule
    \end{tabular}
    }
    \label{tab:sup}
\end{table}

\begin{table}[t]
    \small
    \centering
    \caption{Ablation: comparison of different structural encoding schemes and their combinations.
    }
    % \begin{tabular}{l|ccc|ccc|ccc|ccc} \toprule
    % {Method} & \multicolumn{3}{c}{MUTAG} & \multicolumn{3}{c}{PROTEINS} & \multicolumn{3}{c}{PTC} & \multicolumn{3}{c}{NCI1}  \\
    % & None & Lap & GCKN & None & Lap & GCKN & None & Lap & GCKN & None & Lap & GCKN  \\
    % \midrule
    % % {Size} & {188} & {1113} & {344} & {4110} & {12k} \\
    % % {Classes} & {2} & {2} & {2} & {2} & {Reg.} \\
    % % {Maximum number of nodes} & {28} & {620} & {109} & {111} & {37} \\
    % {T}  & & & & & & & & & & & &  \\
    % {T + Adj} & & & & & & & & & & & &  \\
    % {T + 2-step} & & & & & & & & & & & &   \\
    % {T + 3-step}& & & & & & & & & & & &   \\
    % {T + diffusion} & & & & & & & & & & & &   \\
    % \bottomrule
    % \end{tabular}
    \resizebox{\textwidth}{!}{
    \begin{tabular}{l|c|c|c|c|c|c} \toprule
         \multirow{2}{*}{Dataset} & \multirow{2}{3cm}{Structure encoding in node features} & \multicolumn{5}{c}{Relative positional encoding in attention scores}  \\
         & & T vanilla & T + Adj & T + 2-step & T + 3-step & T + Diffusion   \\ \midrule
         \multirow{3}{*}{MUTAG} & None & 82.2$\pm$6.3  & 87.2$\pm$9.8 &  85.3$\pm$6.9 & 83.3$\pm$6.3 & 82.7$\pm$7.6 \\
         & LapPE & 85.8$\pm$5.9 & 86.0$\pm$4.2 & 84.7$\pm$4.7 & 83.5$\pm$5.2 & 84.2$\pm$7.2 \\
         & GCKN & 84.4$\pm$7.8 & \textbf{90.5$\pm$7.0} & 90.4$\pm$5.8 &  90.0$\pm$6.3 & 90.0$\pm$6.8 \\ \midrule
         \multirow{3}{*}{PROTEINS} & None & 75.6$\pm$4.9 & 72.4$\pm$4.9 & 72.8$\pm$4.5 & \textbf{76.2$\pm$4.4} & 74.6$\pm$4.2\\
         & LapPE & 74.6$\pm$2.7 & 74.7$\pm$5.2 & 75.0$\pm$4.7 & 74.3$\pm$5.3 & 74.7$\pm$5.3\\
         & GCKN & 69.5$\pm$3.8 & 71.1$\pm$6.9 & 72.5$\pm$4.6 & 70.0$\pm$5.1 & 72.4$\pm$4.9 \\ \midrule
         \multirow{3}{*}{PTC} & None & 58.1$\pm$10.5 & 59.9$\pm$5.9 & \textbf{62.0$\pm$9.4} & 61.0$\pm$6.2 & 59.1$\pm$7.4 \\
         & LapPE & 55.6$\pm$5.0 & 57.1$\pm$3.8 & 58.8$\pm$6.6 & 57.1$\pm$5.3 & 57.3$\pm$7.8 \\
         & GCKN & 61.5$\pm$5.8 & 57.9$\pm$4.2 & 58.4$\pm$7.6 & 55.2$\pm$8.8 & 55.9$\pm$8.1 \\ \midrule
         \multirow{3}{*}{NCI1} & None & 70.0$\pm$4.5 & 79.7$\pm$2.0 & 78.0$\pm$1.5 & 77.6$\pm$3.6 & 78.9$\pm$1.6 \\
         & LapPE & 74.6$\pm$1.9 & 78.7$\pm$1.9 & 78.4$\pm$1.3 & 78.7$\pm$1.5 & 77.8$\pm$1.0 \\
         & GCKN & 78.1$\pm$5.1 & \textbf{81.4$\pm$2.2} & 81.0$\pm$1.8 & 81.0$\pm$1.8 & 81.0$\pm$2.0 \\ \midrule
         \multirow{3}{*}{ZINC} & None & 0.696$\pm$0.007 & 0.243$\pm$0.005 & 0.243$\pm$0.010 & 0.244$\pm$0.011 & 0.255$\pm$0.010 \\
         & LapPE & 0.507$\pm$0.003 & \textbf{0.202$\pm$0.011} & 0.227$\pm$0.030 & 0.210$\pm$0.003 & 0.221$\pm$0.038 \\
         & GCKN & 0.274$\pm$0.011 & 0.211$\pm$0.010 & 0.213$\pm$0.016 &  0.203$\pm$0.011 & 0.218$\pm$0.006 \\ \bottomrule
    \end{tabular}
    }
    \label{tab:structural_encoding}
\end{table}

\paragraph{Comparison of GraphiT and baselines methods.}
We show our results in Table~\ref{tab:sup}. For smaller datasets such as MUTAG, PROTEINS or PTC, our Transformer without positional encoding performs reasonably well compared to GNNs, whereas for NCI1 and ZINC, incorporating structural information into the model is key to good performance. On all datasets, GraphiT is able to perform as well as or better than the baseline GNNs. In particular on ZINC, GraphiT outperforms all previous baseline methods by a large margin. For this, it seems that the factor $D^{-1/2}$ in~\eqref{eq:pos_forward} is important, allowing to capture more information about the graph structure. The answer to \textbf{Q1} is therefore positive.

\vs
\paragraph{Comparison of relative position encoding schemes.}
% \begin{itemize}
%     \item Diffusion always outperformed by the others
%     \item T+ Adj ok on MUTAG and NCI + ZINC
%     \item T 2/3 step ok on PTC and PROTEINS + ZINC
% \end{itemize}
Here, we compare our transformer used with different relative positional encoding strategies, including adjacency (1-step RW kernel with $\gamma=1.0$ corresponding to a normalized adjacency matrix $D^{-1/2}A D^{-1/2}$) which is symmetric but not positive semi-definite, 2 and 3-step RW kernel with $\gamma=0.5$ and a diffusion kernel with $\beta=1$.
Unlike the vanilla transformer that works poorly on big datasets including NCI1 and ZINC, keeping all nodes communicate through our diffusion kernel positional encoding can still achieve performances close to encoding methods relying on local communications such as adjacency or p-step kernel encoding. Interestingly, our adjacency encoding, which could be seen as a variant of the neighborhood aggregation of node features used in GAT, is shown to be very effective on many datasets. In general, sparse local positional encoding seems to be useful for our prediction tasks, which tempers the answer to \textbf{Q1}. 

\vs
\paragraph{Comparison of structure encoding schemes in node features.} 
%
% \begin{itemize}
%     \item On the vanilla transformer, using a structurl encpding without positional encoding is almost always beneficial
%     \item GCKn seem to workwhen on big datasets  and but yields a bad perf for PROTEINS
% \end{itemize}
In this paragraph, we compare different ways of injecting graph structures to the vanilla transformer, including Laplacian PE~\cite{dwivedi2021generalization} and unsupervised GCKN-path features~\cite{Chen2020}. We observe that
incorporating topological structures directly into the features of input nodes is a very useful strategy for vanilla transformers. This yields significant performance improvement on almost all datasets except PROTEINS, by either using the Laplacian PE or GCKN-path features. Among them, GCKN features are observed to outperform Laplacian PE by a large margin, except on MUTAG and PROTEINS (third column of Table~\ref{tab:structural_encoding}).
A possible reason for this exception is that PROTEINS seems to be very specific, such that prediction models do not really benefit from encoding positional information. In particular, GCKN brings a pronounced performance boost on ZINC, suggesting the importance of encoding substructures like paths for prediction tasks on molecules.

\vs
\paragraph{Combining relative position encoding and structure encoding in node features.}
Table~\ref{tab:structural_encoding} reports the ablation study of the transformer used with or without structure encoding and coupled or not with a relative positional encoding. The results show that relative PE outperforms the topological Laplacian PE, suggesting a positive answer to \textbf{Q2}. However, using both simultaneously improves the results considerably, especially on ZINC. In fact, combining relative position encoding and a structure encoding scheme globally improves the performances. In particular, using the GCKN-path layer features works remarkably well for all datasets except PROTEINS. More precisely, we see that the combination of GCKN with the adjacent matrix PE yields the best results among the other combinations for MUTAG and NCI1. In addition, the GCKN coupled with the 3-step RW kernel achieves the second best performance for ZINC. The answer to \textbf{Q3} is therefore combining a structure encoding scheme in node features, such as GCKN, with relative positional encoding.
\vs
\paragraph{Discussion.}
% tranafo = pooling
Combining transformer and GCKN features results in substantial improvement over the simple sum or mean global poolings used in original GCKN models on ZINC dataset as shown in Table~\ref{tab:sup}, which suggests that transformers can be considered as a very effective method for aggregating local substructure features on large datasets at the cost of using much more parameters. This point of view has also been explored in~\cite{Mialon2021}, which introduces a different form of attention for sequence and text classification.
%Reducing the computational complexity of transformer while retaining its performance would be an interesting future research direction. 
A potential limitation of GraphiT is its application to large graphs, %especially for our kernel-based relative PE schemes that scale quadratically with the number of nodes (similar to the complexity of self-attention mechanisms for sequences)
as the complexity of the self-attention mechanism scales quadratically with the size of the input sequence. However, a recent line of work coined as efficient transformers alleviated these issues both in terms of memory and computational cost~\citep{Beltagy2020Longformer,kitaev2020reformer}. We refer the reader to the following survey~\citep{tay2020efficient}.

%\begin{itemize}
    %\item Interestingly, simple kernel methods do very well, even without positional encoding. The same goes with a bare Transformers for some data sets (MUTAG, PROTEINS, PTC). This suggests that those data sets are perhaps not the best to evaluate GNNs (although they are popular).
    %\item Unsurprisingly, unsupervised methods do better than most supervised methods for the smaller data sets (MUTAG, PTC).
    %\item The match kernel with Laplacian PE does not converge when we flip the signs. Because the kernel matrix is not p.d anymore? %This suggests that GraphWave is indeed more transferable between graphs.
    %When we don't, the results are worse than without Laplacian PE, except for MUTAG (slightly better). GraphWave is therefore a good position encoding and easy to use (we did not even fine-tuned it).
    %\item OTKE improves for PTC only. It seems that this data set requires capturing larger motifs. We can expect a Transformer to work as well. 
    %\item For smaller datasets (MUTAG, PTC), a fixed positional encoding is more useful. For bigger dataset (PROTEINS, NCI1), it seems that is better to have at least partially learnable position encoding.
    %\item Path Transformers may be prone to overfitting on the smaller datasets as it increases the size of the features.
%\end{itemize}

\subsection{Visualizing attention scores for the Mutagenicity dataset.}
\label{subsec:visu}

We now show that the attention yields visualization mechanisms for detecting important graph motifs.

\begin{figure}
\begin{subfigure}{.2\textwidth}
  \centering
  \includegraphics[width=1.0\linewidth]{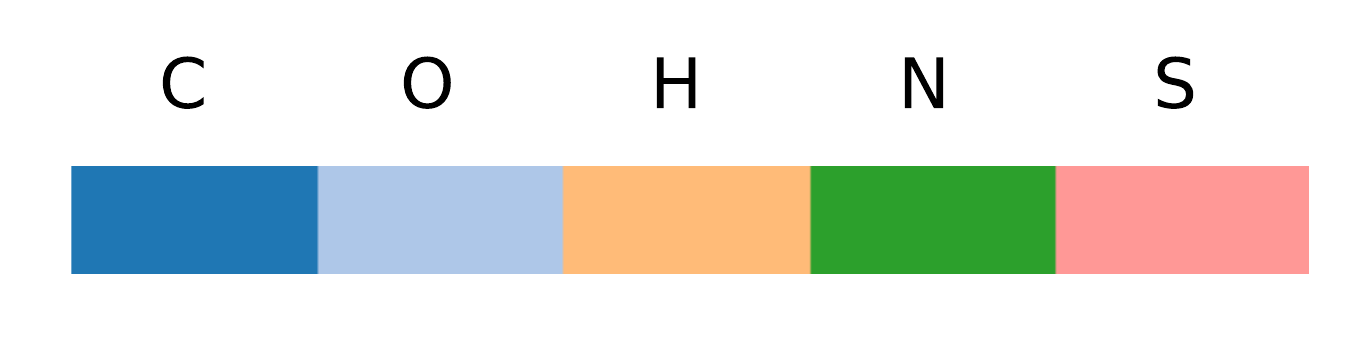}
\end{subfigure}%

\centering
\begin{subfigure}{.5\textwidth}
  \centering
  \includegraphics[width=0.85\linewidth]{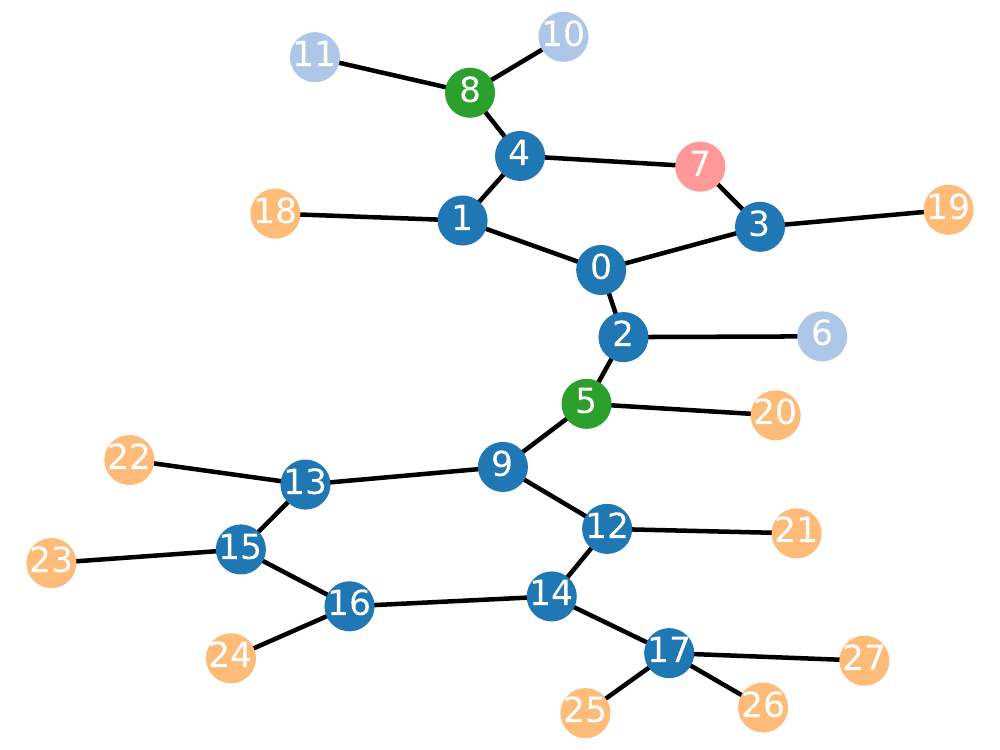}
  \caption{Nitrothiopheneamide-methylbenzene}
  \label{fig:NO2}
\end{subfigure}%
\begin{subfigure}{.5\textwidth}
  \centering
  \includegraphics[width=0.85\linewidth]{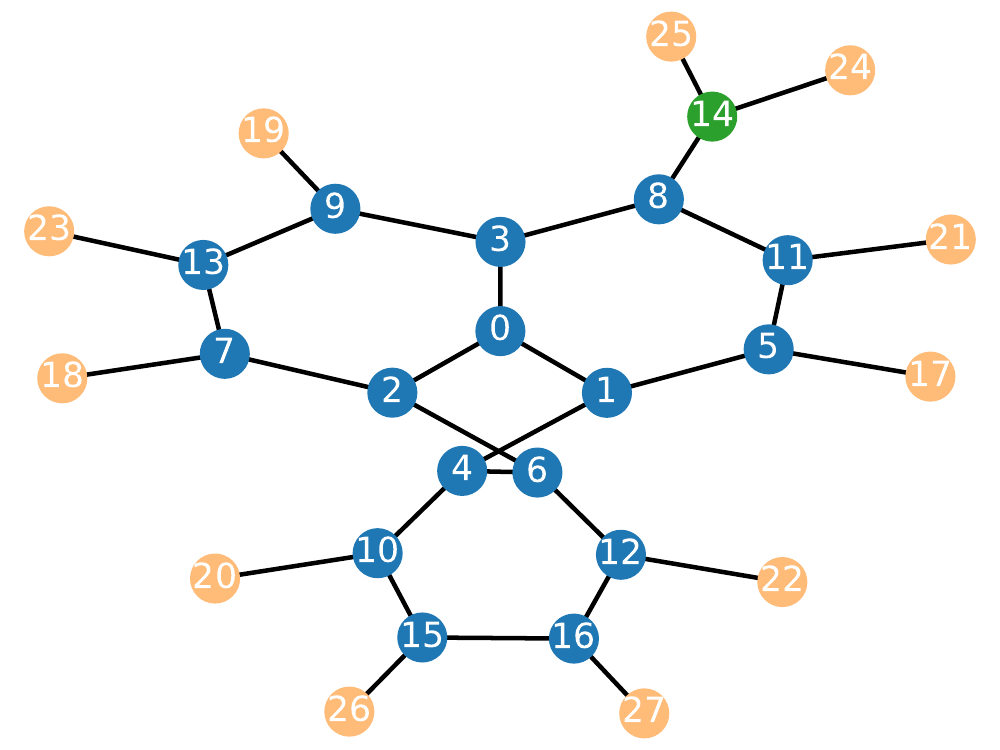}
  \caption{Aminofluoranthene}
  \label{fig:NH2}
\end{subfigure}
\caption{Examples of molecules from Mutagenicity correctly classified as mutagenetic by our model.}
\label{fig:molecules}
\end{figure}

\begin{figure}
\centering
\begin{tabular}{ccc}
\subfloat{\includegraphics[width = 1.6in]{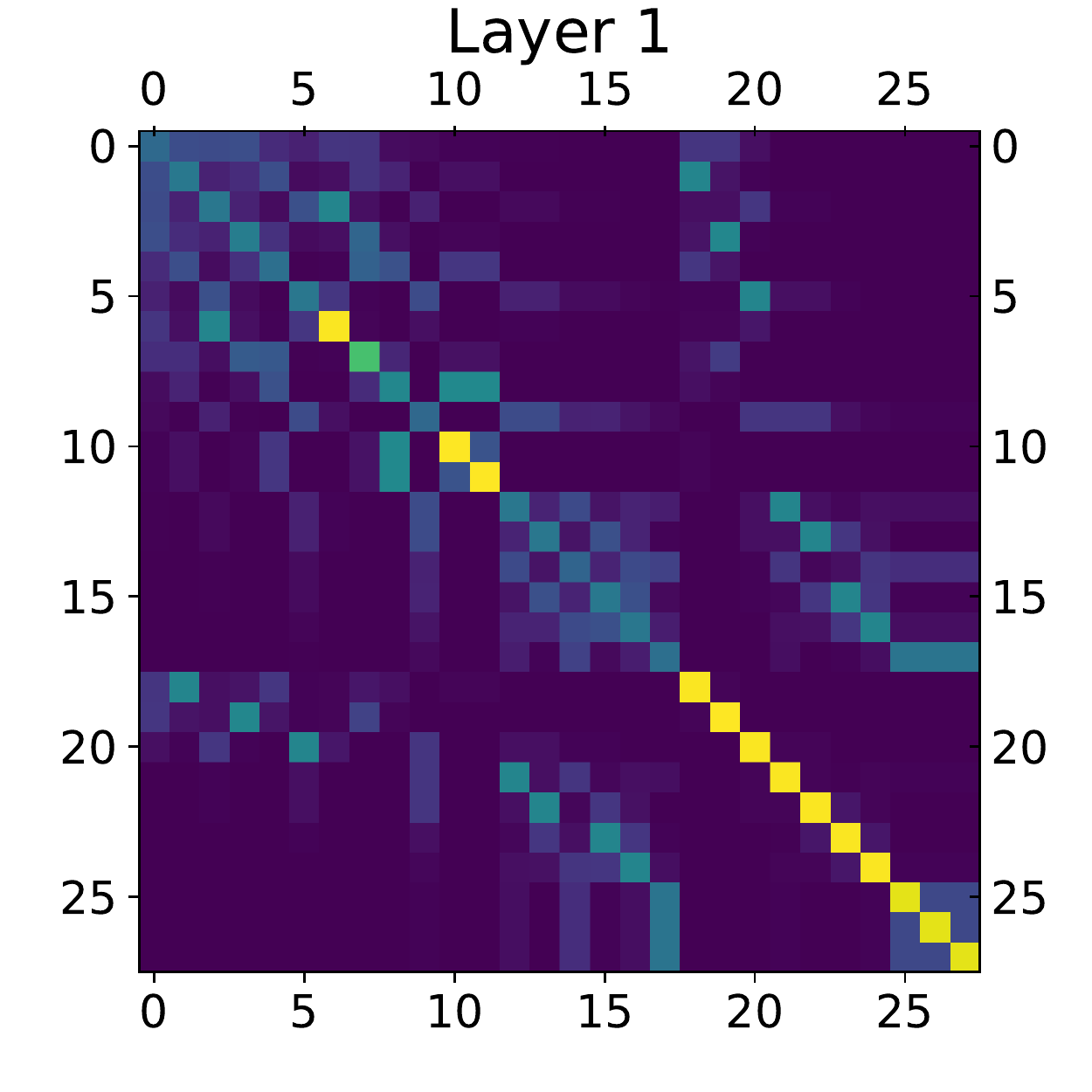}} &
\subfloat{\includegraphics[width = 1.6in]{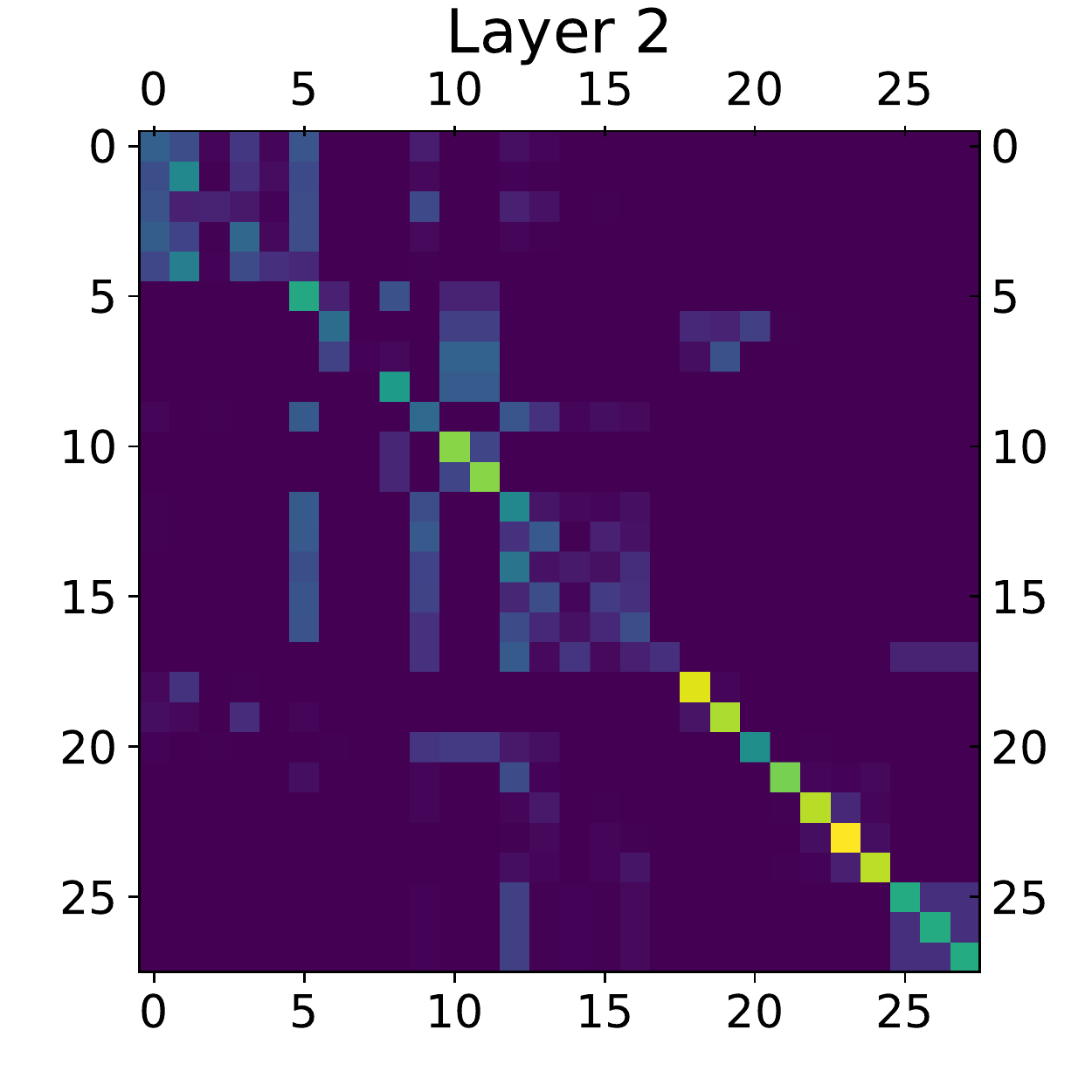}} &
\subfloat{\includegraphics[width = 1.6in]{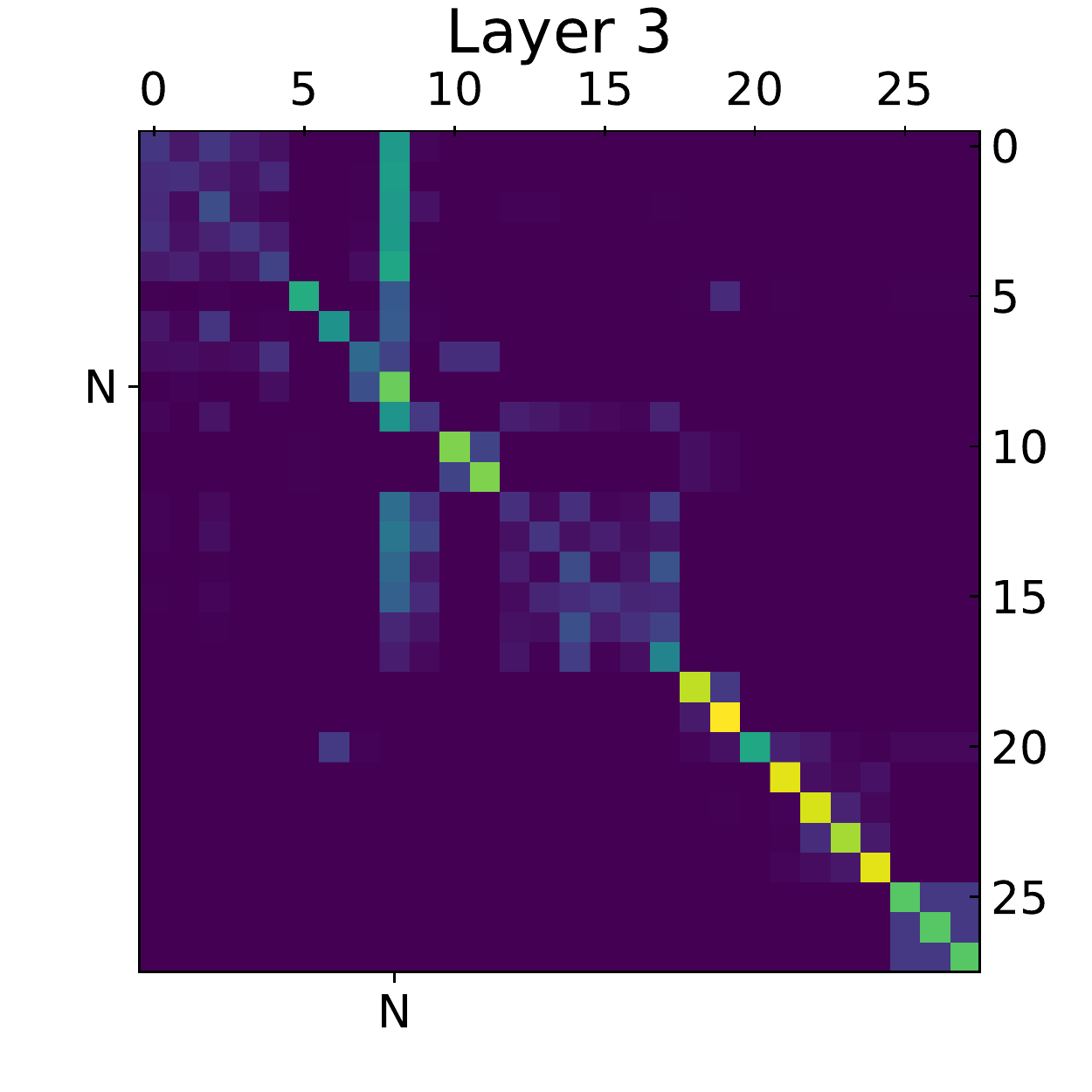}}\\
\subfloat{\includegraphics[width = 1.6in]{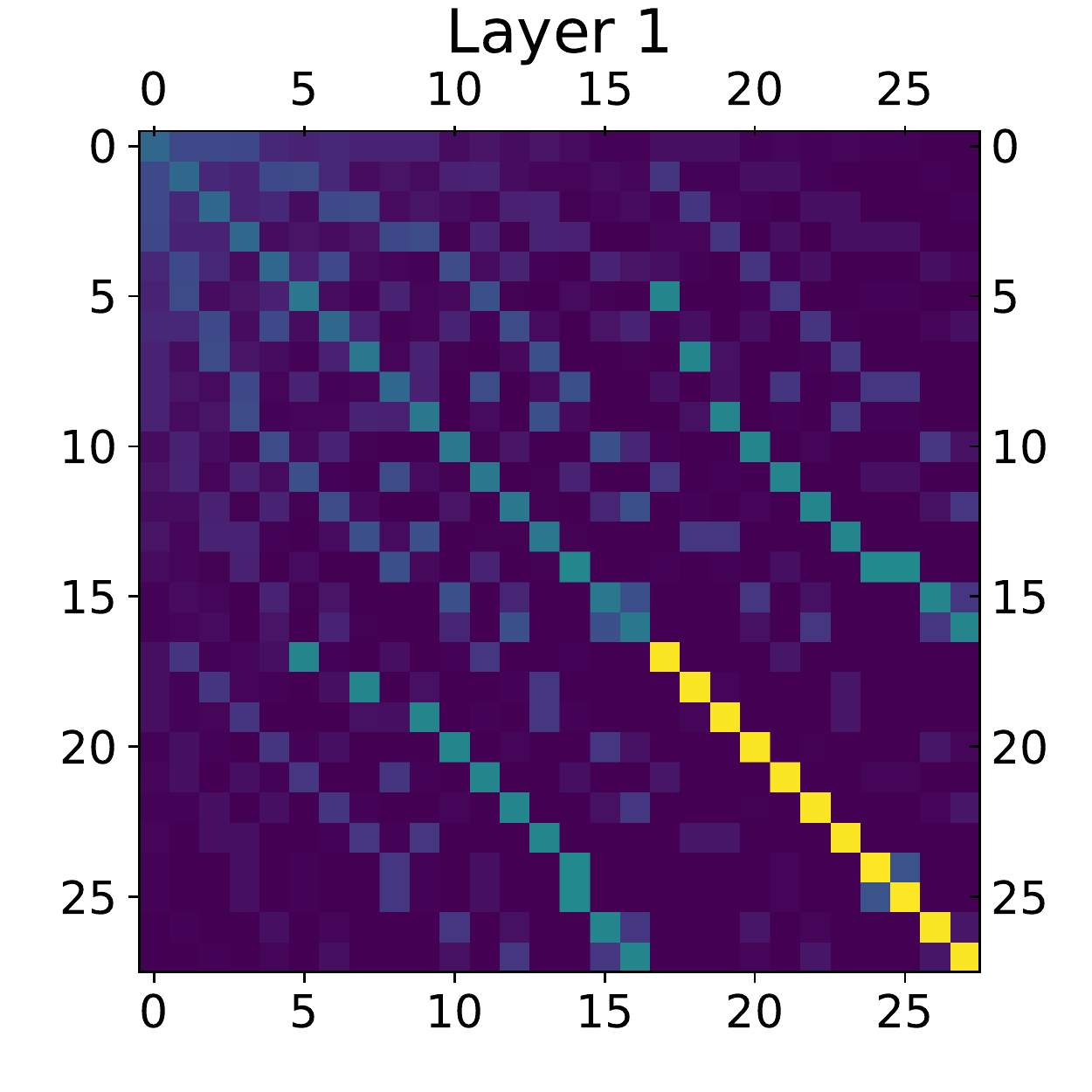}} &
\subfloat{\includegraphics[width = 1.6in]{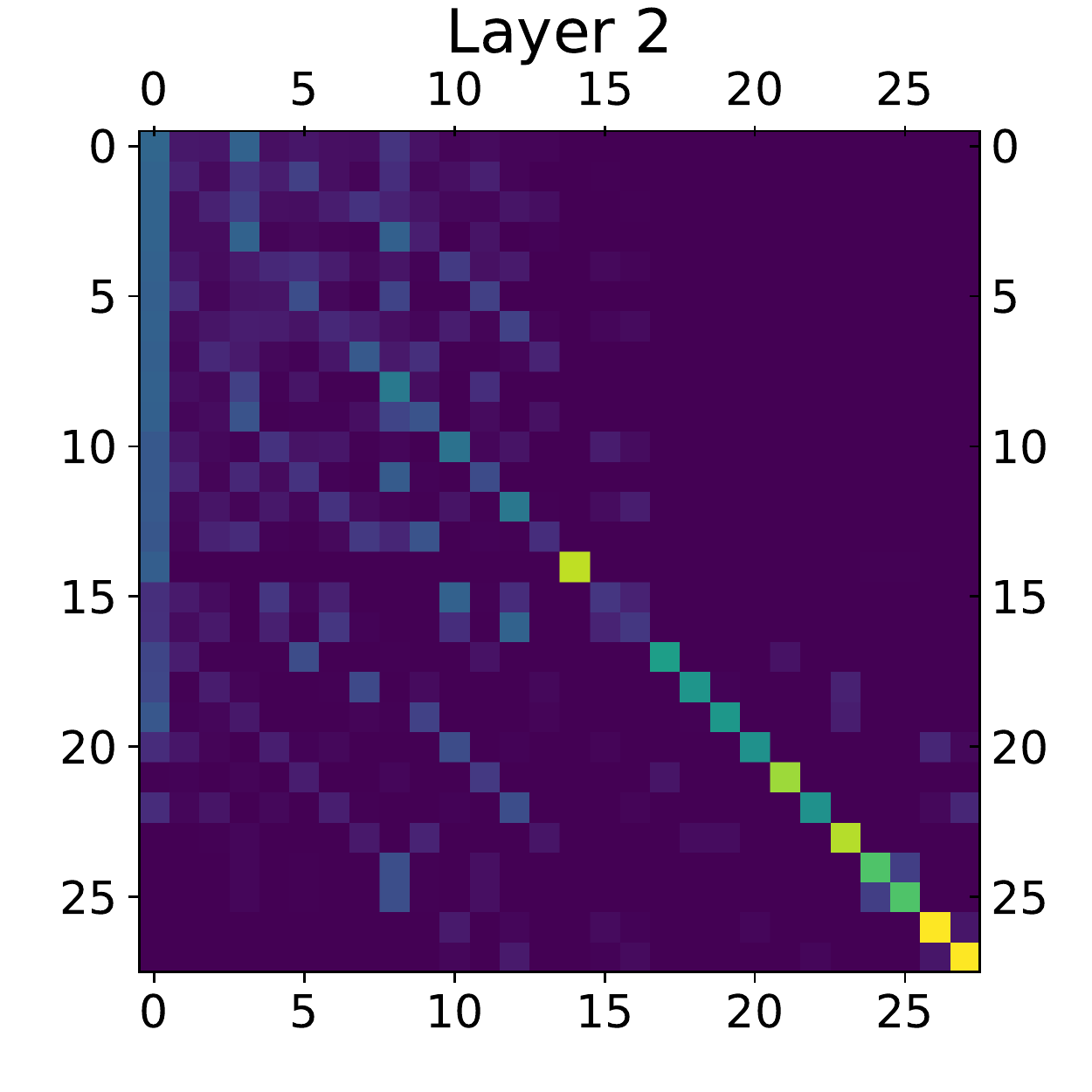}} &
\subfloat{\includegraphics[width = 1.6in]{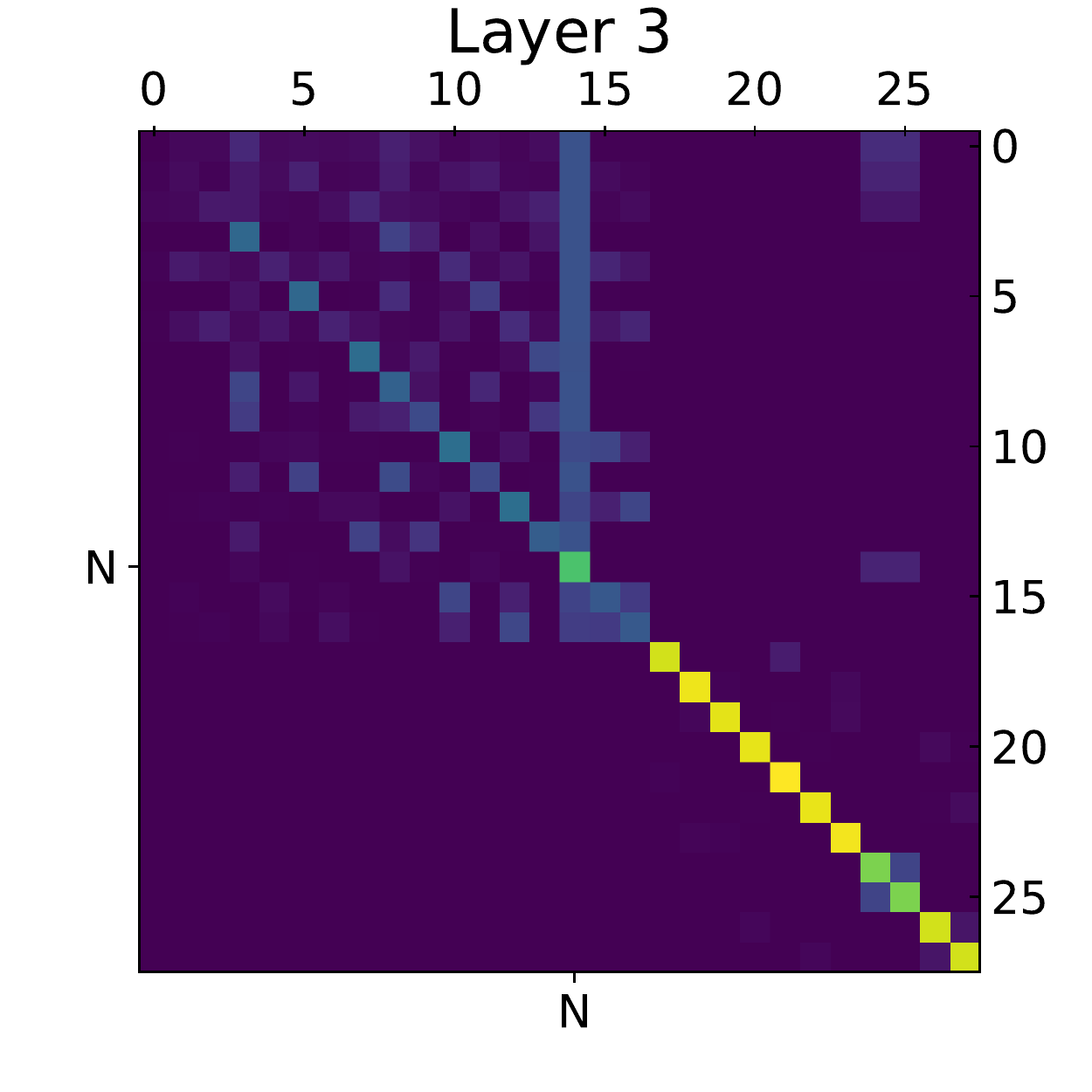}}\\
\end{tabular}
\caption{Attention scores averaged by heads for each layer of our trained model for the compounds in Figures~\ref{fig:NO2} (\textit{Top}) and~\ref{fig:NH2} (\textit{Bottom)}. \textit{Top Left}: diffusion kernel for~\ref{fig:NO2}. %\textit{Top Center}: node $9$ (C of aromatic cycle) is salient.
\textit{Top Right}: node $8$ (N of NO$_2$) is salient. %and $17$ (C of CH$_3$) are salient.
\textit{Bottom Left}: diffusion kernel for~\ref{fig:NH2}. %\textit{Bottom Center}: nodes $3$ (C of fluoranthene) and $8$ (C of aromatic cycle connected to NH$_2$) are salient.
\textit{Bottom Right}: node $14$ (N of NH$_2$) is salient.}
\label{fig:attentions}
\end{figure}

\vs
\paragraph{Mutagenicity.} In chemistry, a mutagen is a compound that causes genetic mutations. This important property is related to the molecular substructures of the compound. The Mutagenicity dataset~\citep{KKMMN2016} contains 4337 molecules to be classified as mutagen or not, and aims at better understanding which substructures cause mutagenicity. We train GraphiT with diffusion position encoding on this dataset with the aim to study whether important substructures can be detected. To this end, we feed the model with molecules of the dataset and collect the attention scores of each layer averaged by heads, as can be seen in Figure~\ref{fig:attentions} for the molecules of Figure~\ref{fig:molecules}. 

\vs
\paragraph{Patterns captured in the attention scores.} While the scores in the first layer are close to the diffusion kernel, the following layers get sparser. Since the attention matrix is multiplied on the right by the values, the coefficients of the node aggregation step for the n-th node is given by the n-th \textit{row} of the attention matrix. As a consequence, salient \textit{columns} suggest the presence of important nodes. After verification, for many samples of Mutagenicity fed to our model, salient atoms indeed represent important groups in chemistry, many of them being known for causing mutagenicity. In compound~\ref{fig:NO2}, the N atom of the nitro group (NO$_2$) is salient. %, the C atom of the methyl group (CH$_3$) %and the C atom connecting the aromatic cycle to the rest of the molecule are salient.
\ref{fig:NO2} was correctly classified as a mutagen by our model and the nitro group is indeed known for its mutagenetic properties~\citep{chung1996}. Note how information flows from O atoms to N in the first two layers and then, at the last layer, how every element of the carbon skeleton look at N, \textit{i.e} the NO$_2$ group. %The H atoms play little to no role in the scores.
We can apply the same reasoning for the amino group (NH$_2$) in compound~\ref{fig:NH2}~\citep{berry1985}. We were also able to identify long-range intramolecular hydrogen bounds such as between H and Cl in other samples. We provide more examples of visualization in Appendix~\ref{sec:add_vis}.
%\begin{figure}
%\centering
%\begin{subfigure}{1.0\textwidth}
%  \centering
%  \includegraphics[width=1.0\linewidth]{figures/attentions.png}
  %\caption{}
%  \label{fig:NO2}
%\end{subfigure}%

%\begin{subfigure}{1.0\textwidth}
 % \centering
 % \includegraphics[width=1.0\linewidth]{figures/attns_NH2.png}
  %\caption{}
 % \label{fig:NH2}
%\end{subfigure}
%\caption{Attention scores averaged by heads for each layer of our trained model for the compound in Figure~\ref{fig:NO2}. \textit{Left}: diffusion kernel. \textit{Center}: node $9$ (C of aromatic cycle) is salient. \textit{Right}: nodes $8$ (N of N$O_2$) and $17$ (C of CH$_3$) are salient.}
%\label{fig:attentions}
%\end{figure}

%\begin{figure}
%   \centering
%   \includegraphics[scale=0.5]{figures/origin10.pdf}
%    \caption{A compound of Mutagenicity correctly classified as mutagenetic by our model.}
%  \label{fig:molecule}
%\end{figure}

%\begin{figure}
%   \centering
%   \includegraphics[scale=0.25]{figures/attentions.png}
%    \caption{Attention scores averaged by heads for each layer of our trained model for the compound in Figure~\ref{fig:NO2}. \textit{Left}: diffusion kernel. \textit{Center}: node $9$ (C of aromatic cycle) is salient. \textit{Right}: nodes $8$ (N of N$O_2$) and $17$ (C of CH$_3$) are salient.}
%   \label{fig:attentions}
%\end{figure}

%\paragraph{Ideas.}

%\begin{itemize}
%    \item One different $\beta$ for each head for capturing different scales.
%    \item Use the Diffusion kernel at the first attention layer only.
%    \item As in relative positional encoding, also add a positional information to the values.
%\end{itemize}

\section{Conclusion}
\label{sec:conclusion}
In this work, we show that using a transformer to aggregate local substructures with appropriate position encoding, GraphiT, is a very effective strategy for graph representation, and that attention scores allow simple model interpretation. One of the reasons for the success of transformers lies in their scaling properties: in language modeling for example, it has been shown that for an equal large amount of data, the transformer's loss follows a faster decreasing power law than Long-Short-Term-Memory networks when the size of the model increases~\citep{kaplan2020scaling}. 
%Do similar conclusions also hold for graph representation with respect to GNNs? 
We believe that an interesting future direction for this work would be to evaluate if GNNs and GraphiT also scale similarly in the context of large self-supervised pre-training, and can achieve a similar success as in natural language processing.

%\paragraph{Broader impact.}
%Learning with graphs opens many perspectives in multiple scientific domains, notably in computational biology for representing proteins. Our paper investigates a particular architecture for representing graphs, and our goal is to improve the quality of predictions made by machine learning for such modalities. The risk of misuse of this research is similar to any methodological paper about graph neural networks. This risk is nevertheless limited at the moment since graph neural networks are less deployed in the real life than convolutional neural networks for images, or classical transformer architectures for natural language processing.

%\begin{itemize}
%    \item Hopefully, pre-training of Transformers at scale. So, energy consumption yet also benefits if some useful pre-trained model is achieved.
%    \item Graph data is varied, social networks, proteins, can be good or bad.
%\end{itemize}

%\newpage

\section*{Acknowledgments}

GM, DC, MS, and JM were supported by the ERC grant number 714381 (SOLARIS project) and by ANR 3IA MIAI@Grenoble Alpes, (ANR-19-P3IA-0003). This work was granted access to the HPC resources of IDRIS under the allocation 2021-AD011012521 made by GENCI. GM thanks Robin Strudel for useful discussions on transformers.

\bibliographystyle{abbrv}
\bibliography{biblio}

\newpage
\appendix

\vspace*{0.3cm}
\begin{center}
   {\huge \textbf{Appendix}}
\end{center}
\vspace*{0.5cm}

\section{Experimental Details}
\label{sec:add_exp}
In this section, we provide implementation details and additional experimental results.

\subsection{General Details.}

\paragraph{Computing infrastructure.} Computations were done on a GPU cluster equipped with Tesla V100-16G and Tesla V100-32G. We have monitored precisely the entire computational cost of this research project (including preliminary experiments, designing early and final models, evaluating baselines, running final experiments), which was approximately 20k GPU hours. 

\paragraph{Position encoding and structure encoding parameters for all datasets.} 
$\gamma$ for p-step random walk kernels is fixed to 0.5 for both 2- and 3-step random walk kernels. $\beta$ for the diffusion kernel is fixed to 1.0. Regarding the structure encoding in node features, the dimension of Laplacian positional encoding is set to 8 for ZINC as suggested by~\cite{dwivedi2021generalization} and to 2 for graph classification datasets. The path size, the filter number and the bandwidth parameter of the unsupervised GCKN path features (used for structure encoding) are set to 8, 32 and 0.6 respectively for the ZINC dataset whereas the path size is set to 5 for graph classification datasets.

\paragraph{Other details.} For all instances of GraphiT, the hidden dimensions of the feed-forward hidden layer in each layer of the encoder are fixed to twice of the dimensions of the attention.

\subsection{Graph Classification Datasets}

Here, we provide experimental details for MUTAG, PROTEINS, PTC, NCI1 and Mutagenicity datasets.

\paragraph{Datasets.} These free datasets were collected by~\cite{KKMMN2016} for academic purpose. MUTAG consists in classifying molecules into mutagenetic or not. PROTEINS consists in classifying proteins into enzymes and non-enzymes. PTC consists in classifying molecules into carcinogenetic or not. NCI1 consists in classifying molecules into positive or negative to cell lung cancer. Mutagenicity consists in classifying compounds into mutagenetic or not.

\paragraph{Training splits.} For MUTAG, PROTEINS, PTC and NCI1, our outer splits correspond to the splits used in~\cite{xu2019powerful}. Our inner splits that divide their train splits into our train and validation splits are provided in our code. The error bars are computed as the standard deviation of the test accuracies across the 10 outer folds. On the other side, as the purpose of using Mutagenicity is model interpretation, we use simple train, validation and test splits respectively composed of $80\%$, $10\%$, and $10\%$ of the whole dataset. 

\paragraph{Hyperparameter choices.} For smaller datasets (MUTAG, PTC, PROTEINS), we use the search grids in Table~\ref{tab:gnn_small} for GNNs, the search grids in Table~\ref{tab:gckn} for GCKN and the search grids in Table~\ref{tab:transfo_small} for both GraphiT and the transformer model of~\cite{dwivedi2021generalization}. For bigger datasets (NCI1), we use the search grid in Table~\ref{tab:gnn_small} for GNNs, the search grids in Table~\ref{tab:gckn} for GCKN and the search grids in Table~\ref{tab:transfo_big} for both GraphiT and the transformer model of~\cite{dwivedi2021generalization}. The best models are selected based on the validation scores that are computed as the average of the validation scores over the last 50 epochs. Then, the selected model is retrained on train plus validation sets and the average of the test scores over the last 50 epochs is reported.
For Mutagenicity, we train a GraphiT model with 3 layers, 4 heads, 64 hidden dimensions. Initial learning rate was set to 1e-3, weight-decay was fixed to 1e-4 and structural information was encoded only via relative position encoding with the diffusion kernel.

\paragraph{Optimization.} We use the cross-entropy loss and Adam optimizer with batch size 128 for GNNs and 32 for both GraphiT and the transformer model of~\cite{dwivedi2021generalization}. For transformers, we do not observe significant improvement using warm-up strategies on these classification tasks. Thus, we simply follow a scheduler that halves the learning rate after every 50 epochs, as for GNNs. All models are trained for 300 epochs.

\begin{table}
    \small
    \centering
    \caption{Parameter grid for GNNs trained on MUTAG, PROTEINS, PTC, NCI1.}
    \begin{tabular}{l|c} \toprule
    {Parameter} & {Grid} \\
    \midrule
    {Layers} & [1, 2, 3] \\
    {Hidden dimension} & [64, 128, 256] \\
    {Global pooling} & [sum, max, mean]\\
    {Learning rate} & [0.1, 0.01, 0.001] \\
    {Weight decay} & [0.1, 0.01, 0.001] \\
    \bottomrule
    \end{tabular}
    \label{tab:gnn_small}
\end{table}

\begin{table}
    \small
    \centering
    \caption{Parameter grid for GCKN trained on MUTAG, PROTEINS, PTC, NCI1.}
    \begin{tabular}{l|c} \toprule
    {Parameter} & {Grid} \\
    \midrule
    {Path size} & [3, 5, 7] \\
    {Hidden dimension} & [16, 32, 64] \\
    {Sigma} & [0.5] \\
    {Global pooling} & [sum, max, mean]\\
    {Learning rate} & [0.001] \\
    {Weight decay} & [0.01, 0.001, 0.0001] \\
    \bottomrule
    \end{tabular}
    \label{tab:gckn}
\end{table}

\begin{table}
    \small
    \centering
    \caption{Parameter grid for transformers trained on MUTAG, PROTEINS, PTC.}
    \begin{tabular}{l|c} \toprule
    {Parameter} & {Grid} \\
    \midrule
    {Layers} & [1, 2, 3] \\
    {Hidden dimension} & [32, 64, 128] \\
    {Heads} & [1, 4, 8]\\
    {Learning rate} & [0.001] \\
    {Weight decay} & [0.01, 0.001, 0.0001] \\
    \bottomrule
    \end{tabular}
    \label{tab:transfo_small}
\end{table}

\begin{table}
    \small
    \centering
    \caption{Parameter grid for transformers trained on NCI1.}
    \begin{tabular}{l|c} \toprule
    {Parameter} & {Grid} \\
    \midrule
    {Layers} & [2, 3, 4] \\
    {Hidden dimension} & [64, 128, 256] \\
    {Heads} & [1, 4, 8]\\
    {Learning rate} & [0.001] \\
    {Weight decay} & [0.01, 0.001, 0.0001] \\
    \bottomrule
    \end{tabular}
    \label{tab:transfo_big}
\end{table}

%\paragraph{Supplementary experiments.} We provide experiments with unsupervised methods on our smallest datasets.

%\begin{table}[h]
%    \small
%    \centering
%    \caption{Mean classification accuracy/regression error on test folds for unsupervised methods. For each test fold, the model was selected by forming a train and validation sets from the remaining folds.
%    }
%    \begin{tabular}{l|c|c|c} \toprule
%    {Method / Dataset} & {MUTAG} & {PROTEINS} & {PTC} \\
%    \midrule
%    {GCKN unsup. (subtree, no aggregation)} & {85.0 $\pm$ 10.2} & {72.8 $\pm$ 5.6} & {61.2 $\pm$ 8.4} \\
%    {GCKN sup. (subtree, aggregation)} & {84.2 $\pm$ 8.0} & {72.8 $\pm$ 4.7} & {58.0 $\pm$ 9.1} \\
%     {Path kernel~\citep{Chen2020}} & {83.3 $\pm$ 10.8} & {74.2 $\pm$ 4.5} & {57.9 $\pm$ 8.2} \\
%    \midrule
    %{GCKN + OTKE unsup. (one layer, agg.)} & {82.8 $\pm$ 8.5} & {72.3 $\pm$ 5.1} & {63.8 $\pm$ 7.2} \\
    %{GCKN + OTKE sup.} & {TODO} & {TODO} & {TODO} \\
    %\midrule
%    {Match kernel (gaussian)} & {82.8 $\pm$ 8.5} & {73.2 $\pm$ 3.6} & {56.5 $\pm$ 10.2} \\
%    {Match kernel (gaussian) +~\citep{dwivedi2021generalization} PE} & {NO CONV.} & {NO CONV.} & {NO CONV.} \\
%    {Match kernel (gaussian) + GraphWave PE} & {86.7 $\pm$ 7.5} & {74.1 $\pm$ 3.9} & {59.1 $\pm$ 10.7} \\
%    \bottomrule
%    \end{tabular}
%    \label{tab:unsup}
%\end{table}

\subsection{Graph Regression Dataset}
Here, we present experimental details and additional experimental results for ZINC dataset.

\paragraph{Dataset.} 

ZINC is a free dataset consisting of 250k compounds and the task is to predict the constrained solubility of the compounds, formulated as a graph property regression problem. This problem is crucial for designing generative models for molecules. Each molecular graph in ZINC has the type of heavy atoms as node features (represented by a binary vector using one-hot encoding) and the type of bonds between atoms as edge features. In order to focus on the exploitation of the topological structures of the graphs, we omitted the edge features in our experiments. They could possibly be incorporated into GraphiT through the approach of~\cite{dwivedi2021generalization} or considering different kernels on graph for each edge bond type, which is left for future work.

\paragraph{Training splits.} 

Following~\cite{dwivedi2021generalization}, we use a subset of the ZINC dataset composed of respectively 10k, 1k and 1k graphs for train, validation and test split. The error bars are computed as the standard deviation of test accuracies across 4 different runs.

\paragraph{Hyperparameter choice.}

In order to make fair comparisons with the most relevant work~\cite{dwivedi2021generalization}, we use the same number of layers, number of heads and hidden dimensions, namely 10, 8 and 64. The number of model parameters for our transformers is only $2/3$ of that of~\cite{dwivedi2021generalization} as we use a symmetric variant for the attention score function in~\eqref{eq:pos_attention}. Regarding the GNNs, we use the values reported in~\cite{dwivedi2021generalization} for GIN, GAT and GCN. In addition, we use the same grid to train the MF model~\citep{duvenaud2015convolutional}, \textit{i.e.}, a learning rate starting at 0.001, two numbers of layers (4 and 16) and the hidden dimension equal to 145. Regarding the GCKN-subtree model, we use the same hidden dimension as GNNs, that is 145. We fix the bandwidth parameter to 0.5 and the path size is fixed to 10. We select the global pooling form max, mean, sum and weight decay from 0.001, 0.0001 and 1e-5, similar to the search grid used in~\cite{dwivedi2021generalization}.

\paragraph{Optimization.} 

Following~\cite{dwivedi2021generalization}, we use the L1 loss and the Adam optimization method with batch size of 128 for training. Regarding the scheduling of the learning rate, we observe that a standard warm-up strategy used in~\cite{Vaswani2017} leads to more stable convergence for larger models (hidden dimension equal to 128). We therefore adopt this strategy throughout the experiments with a warm-up of 2000 iteration steps. 

\paragraph{Additional results.}
Besides the relatively small models presented in Table~\ref{tab:sup}, we also show in Table~\ref{tab:zinc} the performance of larger models with 128 hidden dimensions. Increasing the number of hidden dimensions generally results in boosting performance, especially for transformer variants combined with Laplacian positional encoding in node features. While sparse local positional encoding is shown to be more useful compared to the positional encoding with a diffusion kernel, we show here that a variant of diffusion kernel positional encoding outperforms all other sparse local positional encoding schemes. Since the skip-connection in transformers already assigns some weight to each node itself and the diagonal of our kernels on graphs also have important weight, we considered setting these diagonal to zero in order to remove the effect of the attention scores on self-loop. This modification leads to considerable performance improvement on longer range relative positional encoding schemes, especially for the transformer with diffusion kernel positional encoding combined with GCKN in node features, which results in best performance.

\begin{table}
    % \small
    \centering
    \caption{Mean absolute error for regression problem ZINC. The results are computed from 4 different runs following~\citep{dwivedi2021generalization}.}
    \label{tab:zinc}
    \begin{tabular}{l|c|c|c}\toprule
        % Relative PE in attention score & \multicolumn{3}{c}{ZINC (without edge features)} \\
        \multirow{2}{*}{Relative PE in attention score} & \multicolumn{3}{c}{Structural encoding in node features} \\
         & None & LapPE & GCKN (p=8,d=32) \\ \midrule
        \cite{dwivedi2021generalization} &  {0.3585$\pm$0.0144} & 0.3233$\pm$0.0126  & - \\ \midrule
        Transformers with d=64 \\ \midrule
        T & {0.6964$\pm$0.0067} & {0.5065$\pm$0.0025} & {0.2741$\pm$0.0112} \\
        {T + Adj PE} & 0.2432$\pm$0.0046 & \textbf{0.2020$\pm$0.0112} & {0.2106$\pm$0.0104} \\
        % {T + 1-step RW kernel} & 0.3389$\pm$0.1278 & 0.2145$\pm$0.0051 \\
        {T + 2-step RW kernel} & 0.2429$\pm$0.0096 & 0.2272$\pm$0.0303 & 0.2133$\pm$0.0161  \\
        {T + 3-step RW kernel} & 0.2435$\pm$0.0111 & 0.2099$\pm$0.0027 & 0.2028$\pm$0.0113  \\
        T + diffusion & 0.2548$\pm$0.0102 %(0.2451$\pm$0.0096) 
        & 0.2209$\pm$0.0038 & 0.2180$\pm$0.0055 \\ \midrule
        Setting diagonal to zero, d=64 \\ \midrule
        T & 0.7041$\pm$0.0044 & 0.5123$\pm$0.0232 & 0.2735$\pm$0.0046 \\
        T + 2-step & 0.2427$\pm$0.0053 & 0.2108$\pm$0.0072 & 0.2176$\pm$0.0430 \\
        T + 3-step & 0.2451$\pm$0.0043 & 0.2054$\pm$0.0072 & 0.1986$\pm$0.0091 \\
        T + diffusion & 0.2468$\pm$0.0061 & 0.2027$\pm$0.0084 &  \textbf{0.1967$\pm$0.0023} \\
        \midrule
        Larger models with d=128 \\ \midrule
        T & 0.7044$\pm$0.0061 & 0.4965$\pm$0.0338 & 0.2776$\pm$0.0084 \\
        {T + Adj PE} & 0.2310$\pm$0.0072 & \textbf{0.1911$\pm$0.0094} & 0.2055$\pm$0.0062 \\
        {T + 2-step RW kernel} & 0.2759$\pm$0.0735 
        & 0.2005$\pm$0.0064 & 0.2136$\pm$0.0062 \\
        {T + 3-step RW kernel} & 0.2501$\pm$0.0328 
        & 0.2044$\pm$0.0058 & 0.2128$\pm$0.0069 \\
        T + diffusion & 0.2371$\pm$0.0040 
        & 0.2116$\pm$0.0103 & 0.2238$\pm$0.0068  \\ \midrule
        Setting diagonal to zero, d=128 \\ \midrule
        T & 0.7044$\pm$0.0061 & 0.4964$\pm$0.0340 & 0.2721$\pm$0.0099 \\
        T + 2-step & 0.2348$\pm$0.0010 & 0.2012$\pm$0.0038 & 0.2031$\pm$0.0083 \\
        T + 3-step & 0.2402$\pm$0.0056 & 0.2031$\pm$0.0076 & 0.2019$\pm$0.0084 \\
        T + diffusion & 0.2351$\pm$0.0121 & \textbf{0.1985$\pm$0.0032} & 0.2019$\pm$0.0018 \\ 
        \bottomrule
    \end{tabular}
\end{table}

\section{Additional visualization of Mutagenicity compounds}
\label{sec:add_vis}

In this section, we use attention scores as a visualization mechanism for detecting substructures that may possibly induce mutagenicity. We apply our model to samples from the Mutagenicity data set~\cite{KKMMN2016} and analyze molecules that were correctly classified as mutagenic.

\begin{figure}
\centering
\begin{subfigure}{1.0\textwidth}
\centering
\includegraphics[scale=0.35]{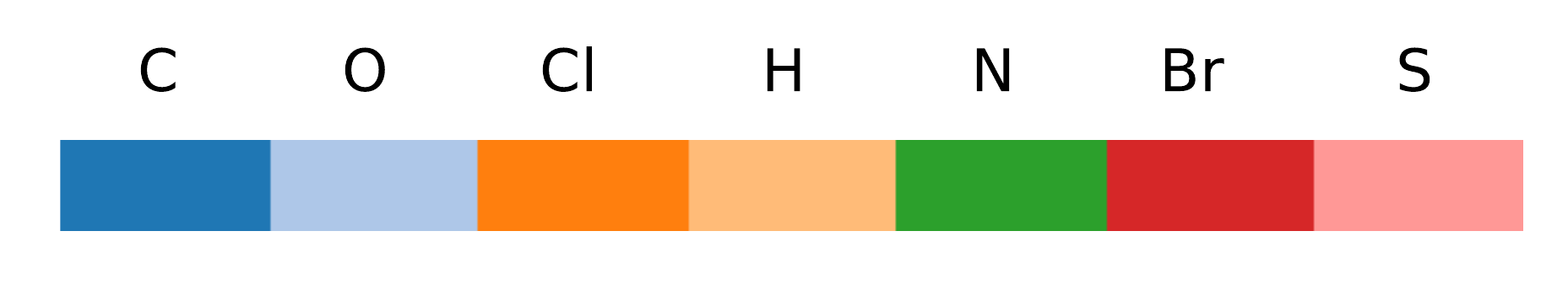}
\end{subfigure}

\vspace{1cm}

\begin{subfigure}{1.0\textwidth}
\centering
\includegraphics[scale=0.8]{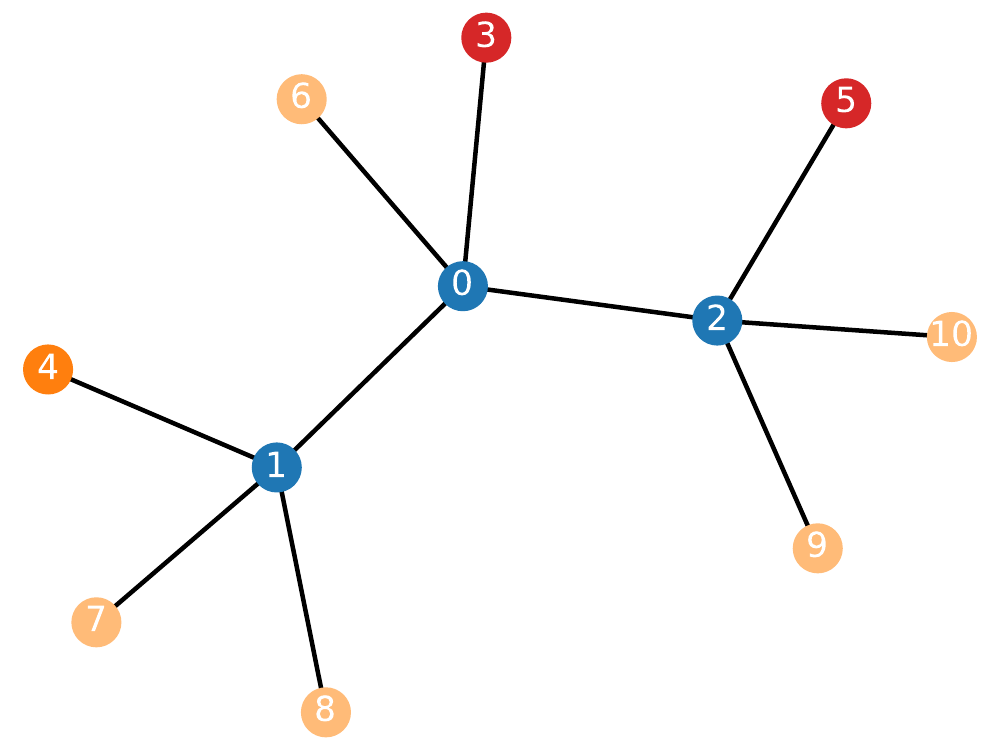}
\end{subfigure}
\caption{1,2-Dibromo-3-Chloropropane.}
\label{fig:dbcp}
\end{figure}

\begin{figure}
\centering
\begin{tabular}{ccc}
\subfloat{\includegraphics[width = 1.6in]{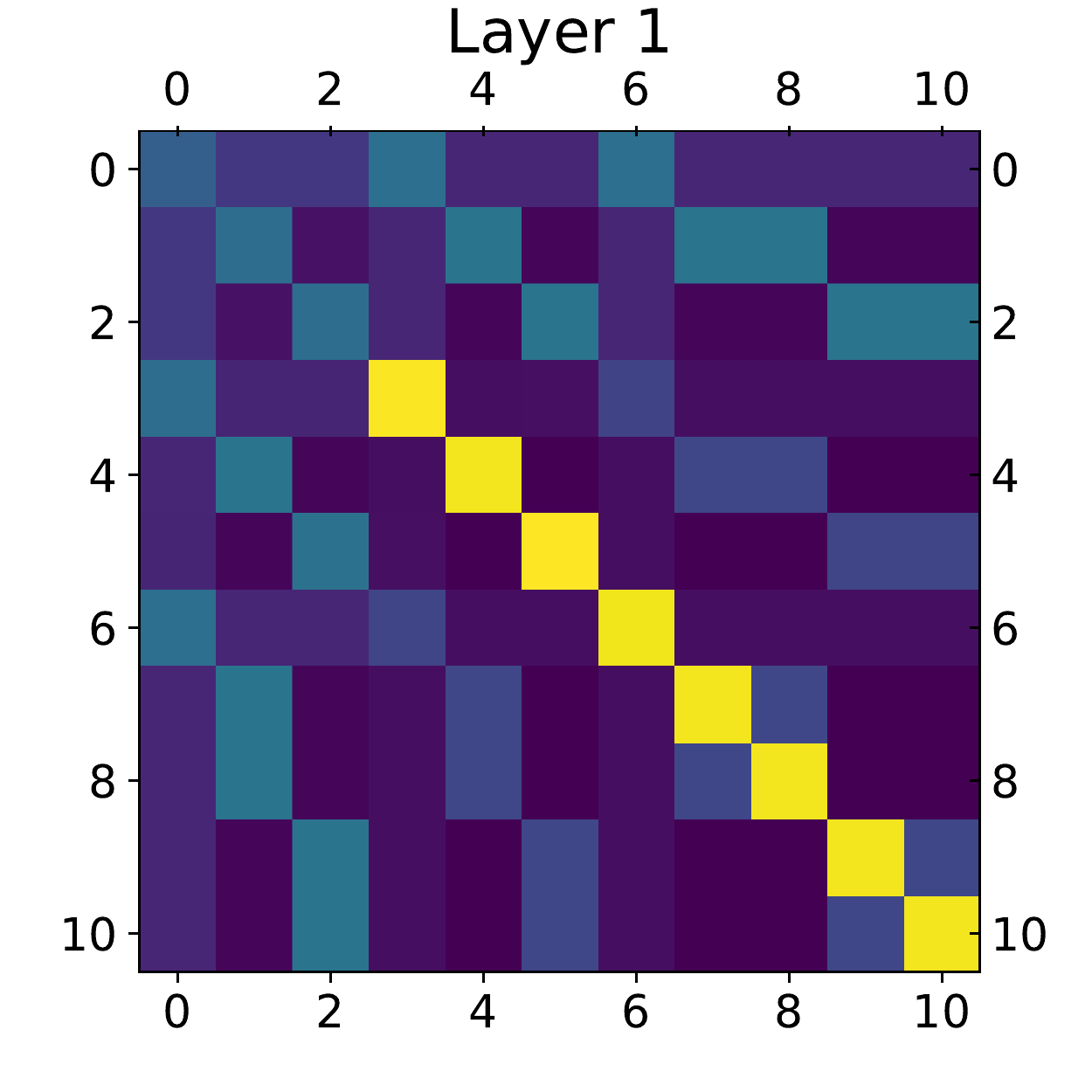}} &
\subfloat{\includegraphics[width = 1.6in]{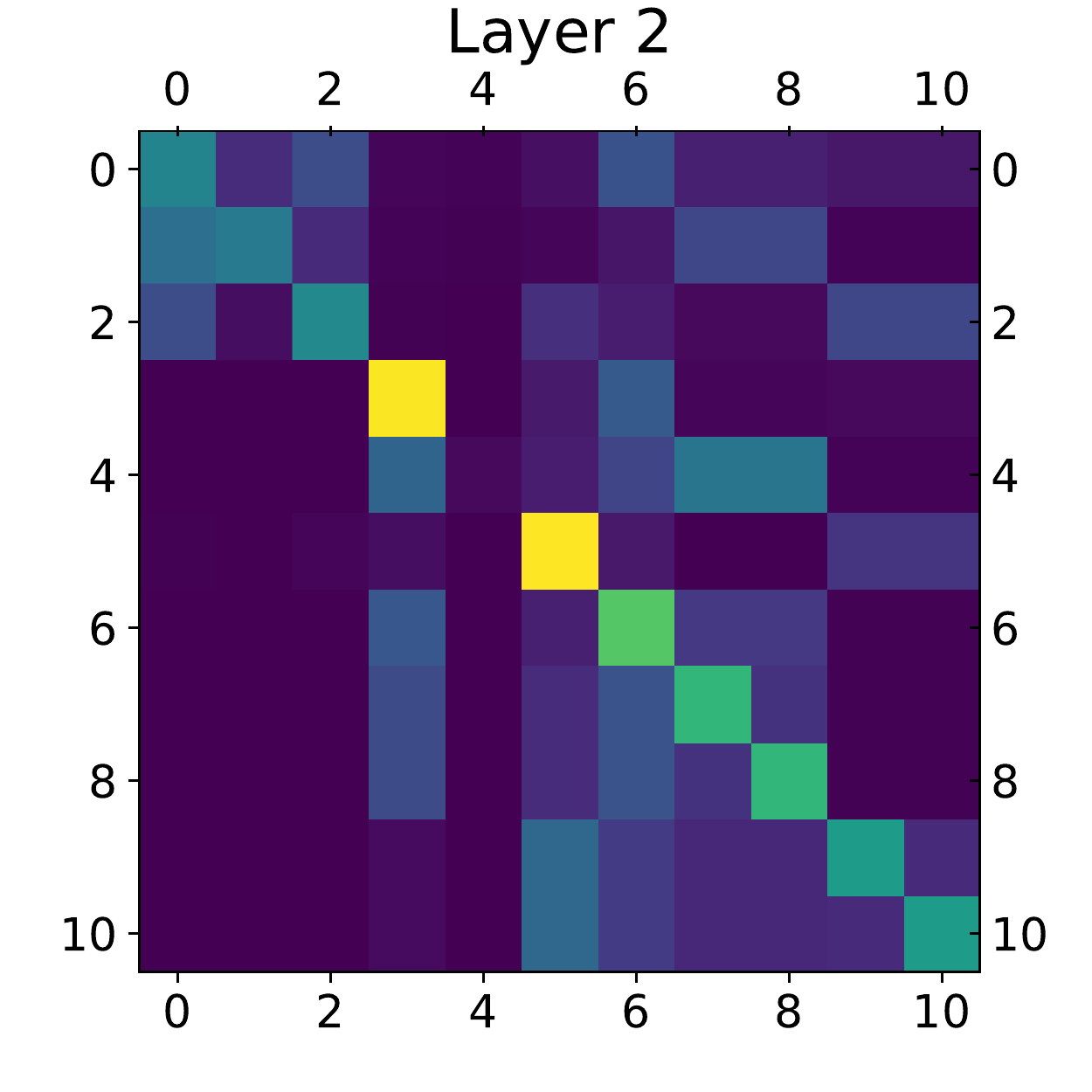}} &
\subfloat{\includegraphics[width = 1.6in]{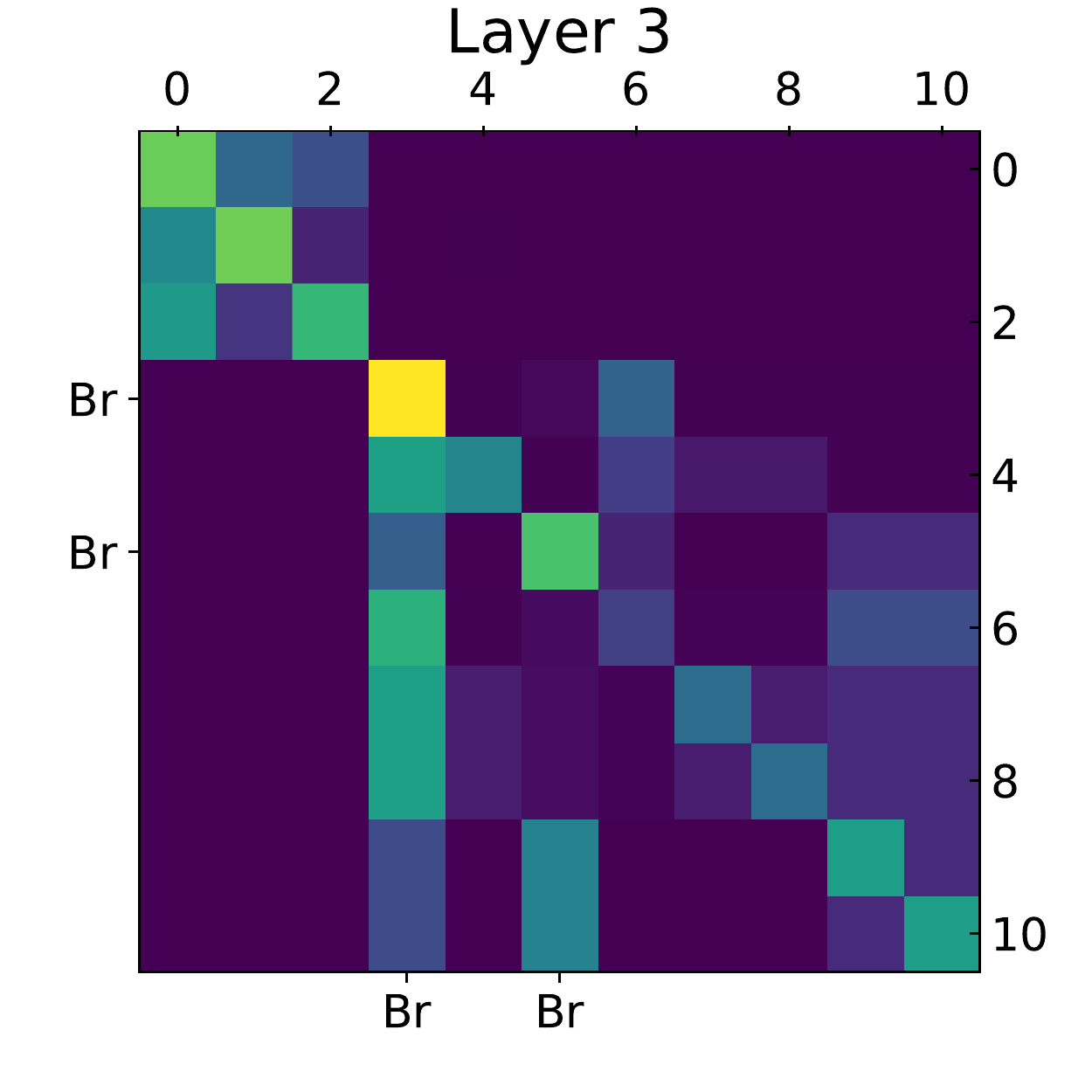}}\\
\end{tabular}
\caption{Attention scores averaged by heads for each layer of our trained model for the compound in Figure~\ref{fig:dbcp}. \textit{Left}: diffusion kernel for~\ref{fig:dbcp}.  \textit{Right}: node $3$ and $5$ (Br) are salient.}
\label{fig:attentions_dbcp}
\end{figure}

\paragraph{1,2-Dibromo-3-Chloropropane (DBCP).} 
DBCP in Figure~\ref{fig:dbcp} was used as a soil fumigant in various countries. It was banned from use in the United Stated in 1979 after evidences that DBCP causes diseases, male sterility, or blindness, which can be instances of mutagenicity. In Figure~\ref{fig:attentions_dbcp}, attention focuses on the carbon skeleton and on the two Bromine (Br) atoms, the latter being indeed known for being associated with mutagenicity~\cite{LAG199473}.

%\begin{figure}
%\centering
%\includegraphics[scale=0.8]{figures/origin2012.pdf}
%\caption{1,2-Dibromo-3-Chloropropane.}
%\label{fig:dbcp}
%\end{figure}

\begin{figure}
\centering
\begin{subfigure}{1.0\textwidth}
\centering
\includegraphics[scale=0.35]{colorbar_appendix.pdf}
\end{subfigure}

\vspace{1cm}

\begin{subfigure}{1.0\textwidth}
\centering
\includegraphics[scale=0.8]{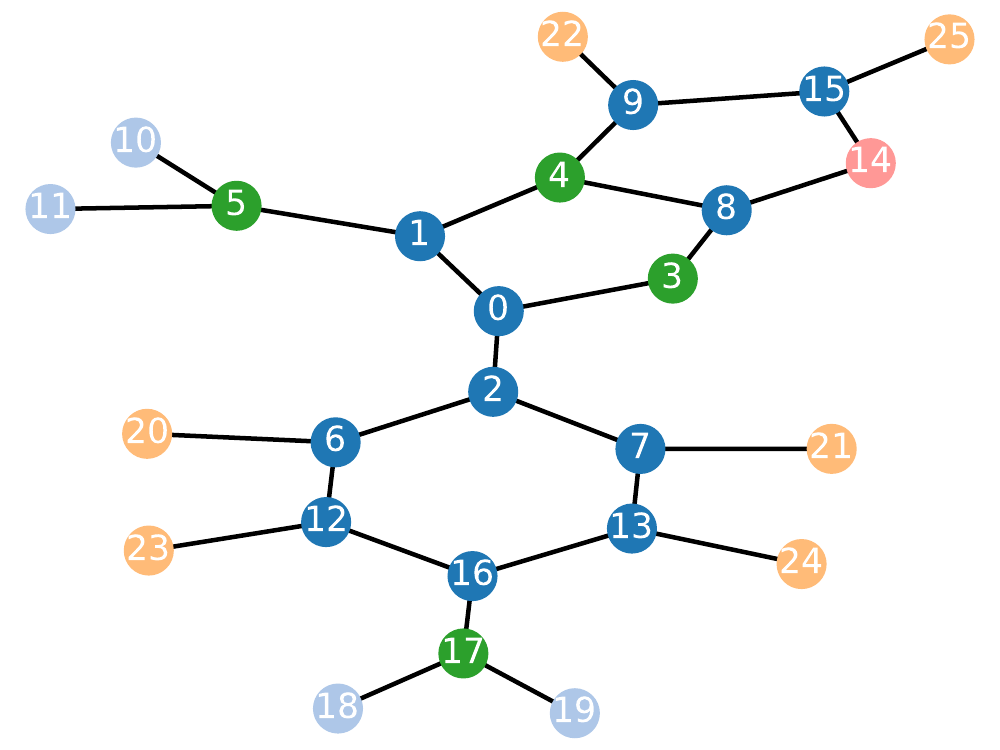}
\end{subfigure}
\caption{Nitrobenzene-nitroimidazothiazole.}
\label{fig:2NO2}
\end{figure}

\paragraph{Nitrobenzene-nitroimidazothiazole.} This compound is shown in Figure~\ref{fig:2NO2}. As for compound~\ref{fig:NO2} in Section~\ref{subsec:visu}, our model puts emphasis on two nitro groups which are indeed known for inducing mutagenicity.

%\begin{figure}
%\centering
%\includegraphics[scale=0.8]{figures/origin2010.pdf}
%\caption{Nitrobenzene-nitro-hexahydroimidazo-thiazole.}
%\label{fig:2NO2}
%\end{figure}

\begin{figure}
\centering
\begin{tabular}{ccc}
\subfloat{\includegraphics[width = 1.6in]{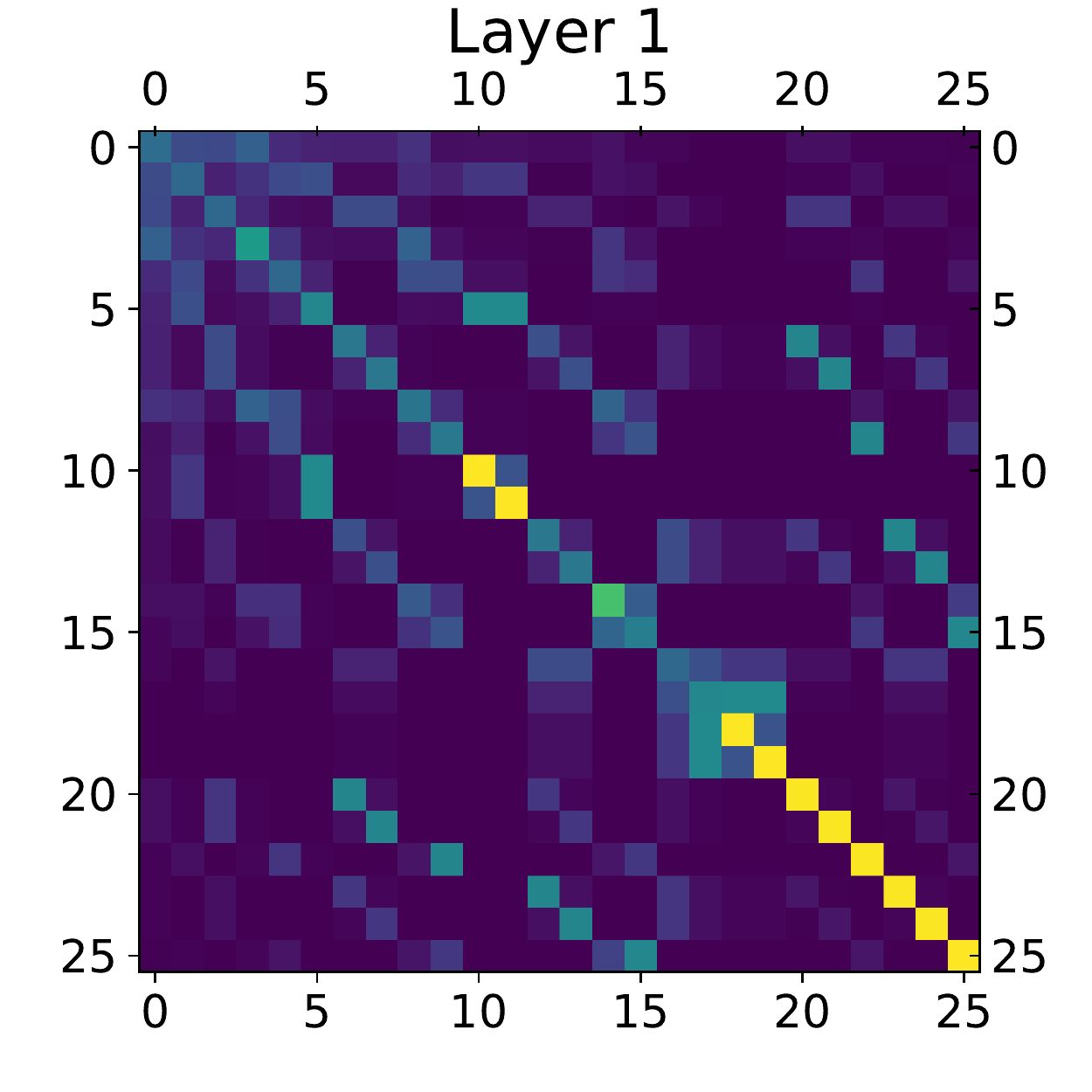}} &
\subfloat{\includegraphics[width = 1.6in]{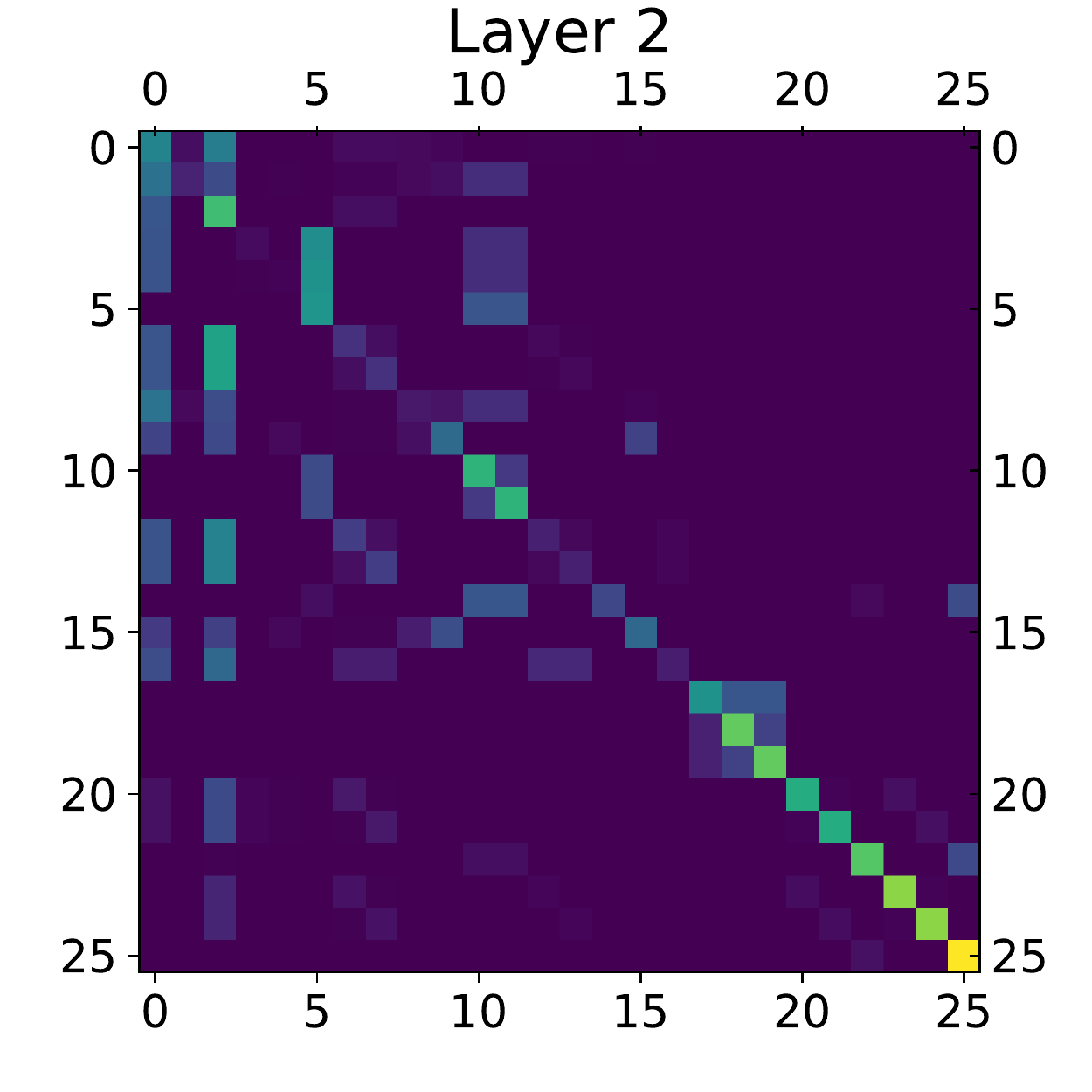}} &
\subfloat{\includegraphics[width = 1.6in]{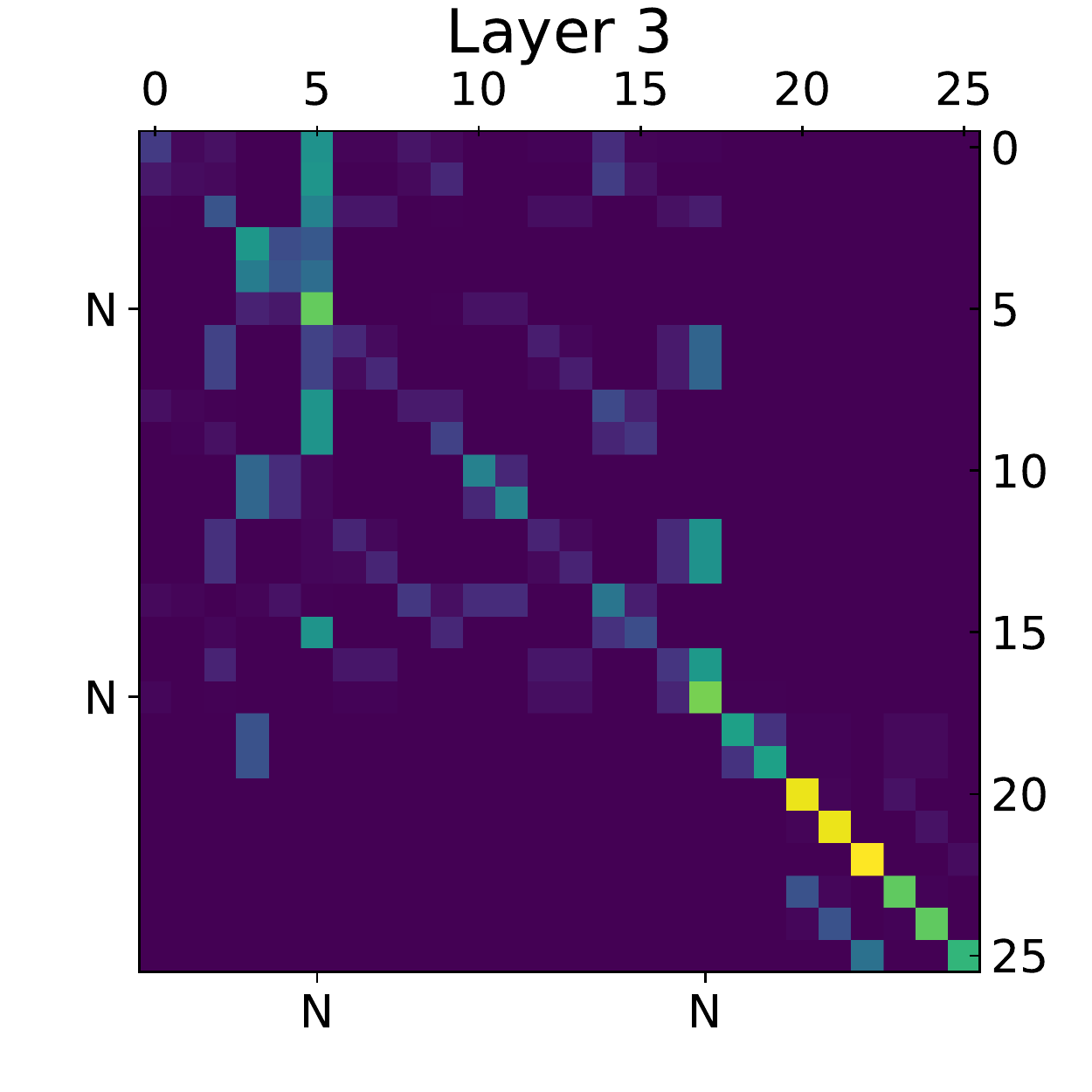}}\\
\end{tabular}
\caption{Attention scores averaged by heads for each layer of our trained model for the compound in Figure~\ref{fig:2NO2}. \textit{Left}: diffusion kernel for~\ref{fig:2NO2}.  \textit{Right}: node $5$ and $17$ (N) are salient.}
\label{fig:attentions_2NO2}
\end{figure}

%\paragraph{Chloroethylcarbamoylesulfide-purine.} Our model puts emphasis on remarkable atoms: Sulfur (S), Nitrogen (N), and Chlorine (Cl). To the best of our knowledge in chemistry, their role as mutagenicity-inducing groups is unclear for this compound, perhaps suggesting the interest of further studying the influence of these groups.

%\begin{figure}
%\centering
%\begin{subfigure}{1.0\textwidth}
%\centering
%\includegraphics[scale=0.35]{figures/colorbar_appendix.pdf}
%\end{subfigure}

%\vspace{1cm}

%\begin{subfigure}{1.0\textwidth}
%\centering
%\includegraphics[scale=0.8]{figures/origin2005.pdf}
%\end{subfigure}
%\caption{Chloroethylcarbamoylesulfide-purine.}
%\label{fig:purine}
%\end{figure}

%\begin{figure}
%\centering
%\includegraphics[scale=0.8]{figures/origin2005.pdf}
%\caption{Chloroethylcarbamoyd-sulfur-purine.}
%\label{fig:purine}
%\end{figure}

%\begin{figure}
%\centering
%\begin{tabular}{ccc}
%\subfloat{\includegraphics[width = 1.6in]{figures/attns_2005_1_t.pdf}} &
%\subfloat{\includegraphics[width = 1.6in]{figures/attns_2005_2_t.pdf}} &
%\subfloat{\includegraphics[width = 1.6in]{figures/attns_2005_3_t.pdf}}\\
%\end{tabular}
%\caption{Attention scores averaged by heads for each layer of our trained model for the compound in Figure~\ref{fig:purine}. \textit{Left}: diffusion kernel for~\ref{fig:purine}.  \textit{Right}: node $3$ and $5$ (Br) are salient.}
%\label{fig:attentions_purine}
%\end{figure}

\begin{figure}
\centering
\begin{subfigure}{1.0\textwidth}
\centering
\includegraphics[scale=0.35]{colorbar_appendix.pdf}
\end{subfigure}

\vspace{1cm}

\begin{subfigure}{1.0\textwidth}
\centering
\includegraphics[scale=0.8]{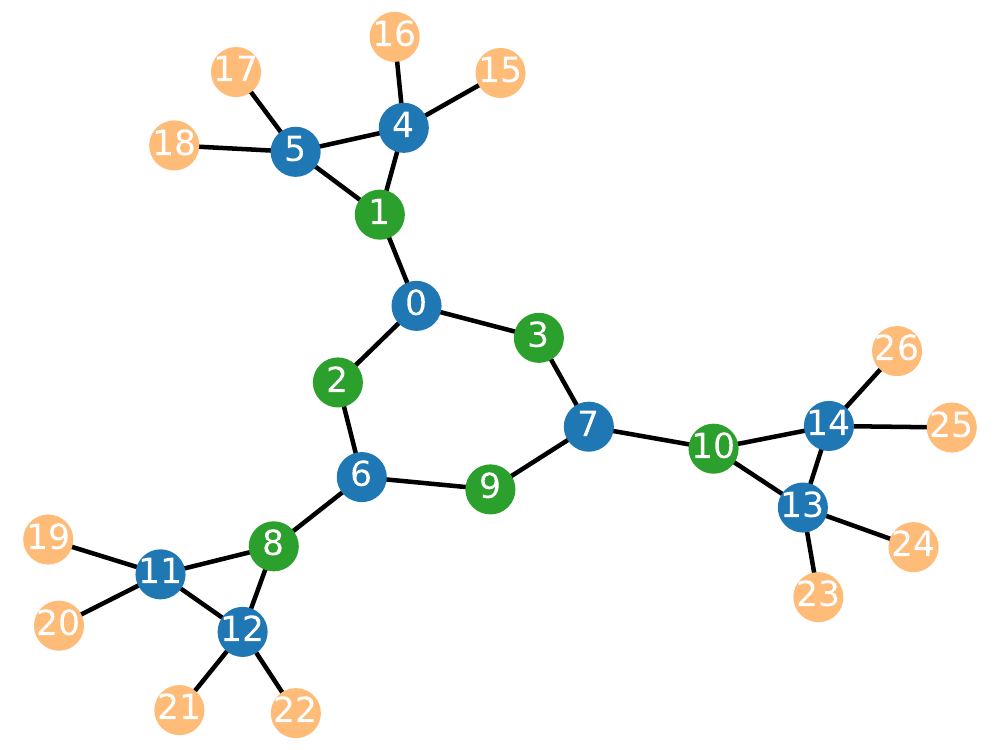}
\end{subfigure}
\caption{Triethylenemelamine.}
\label{fig:tmel}
\end{figure}

\paragraph{Triethylenemelamine.} Triethylenemelamine in Figure~\ref{fig:tmel} is a compound exhibiting mutagenic properties, and is used to induce cancer in experimental animal models. Our model focuses on the three nitrogen atoms of the aziridine groups which are themselves mutagenic compounds.

%\begin{figure}
%\centering
%\includegraphics[scale=0.8]{figures/origin2005.pdf}
%\caption{Chloroethylcarbamoyd-sulfur-purine.}
%\label{fig:purine}
%\end{figure}

\begin{figure}
\centering
\begin{tabular}{ccc}
\subfloat{\includegraphics[width = 1.6in]{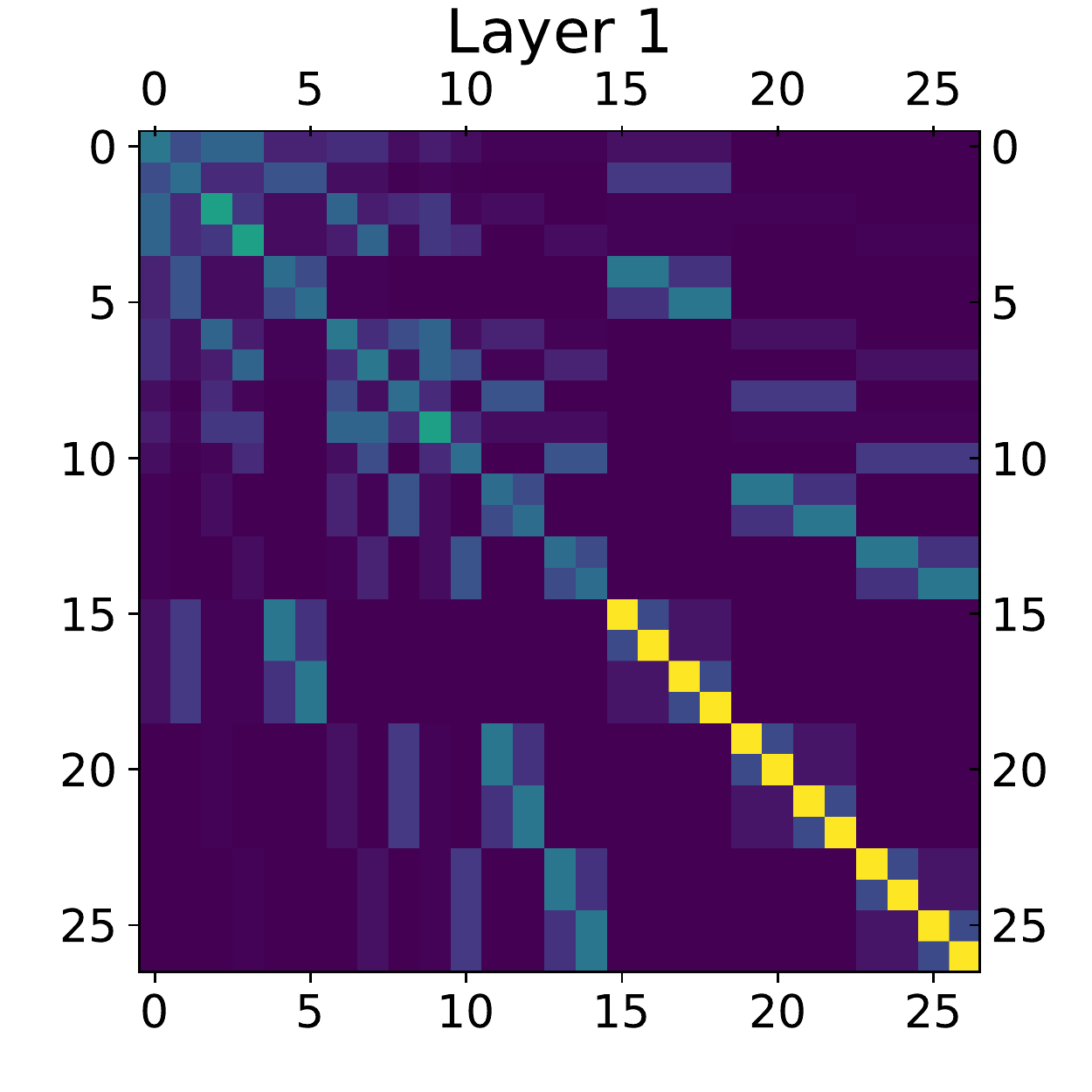}} &
\subfloat{\includegraphics[width = 1.6in]{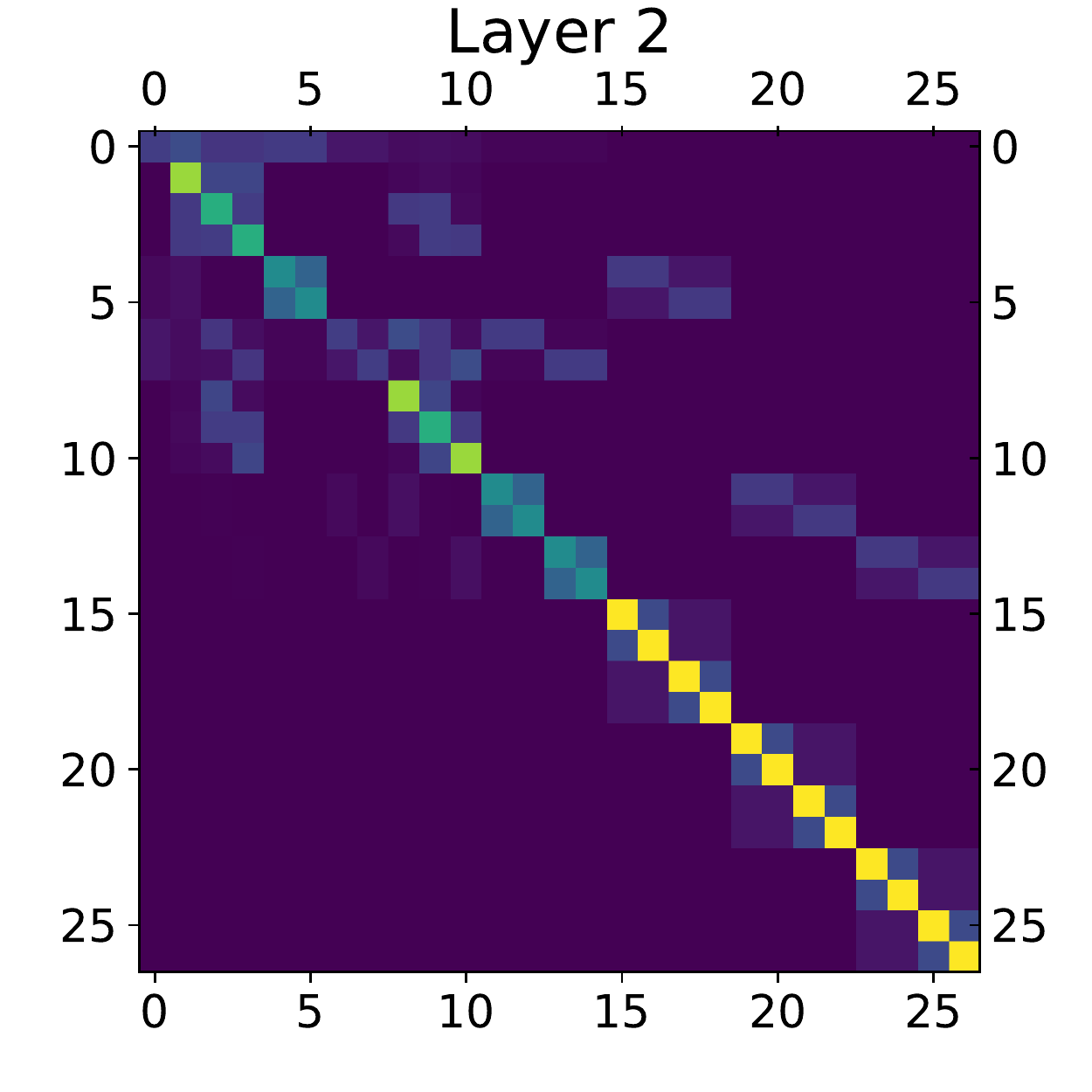}} &
\subfloat{\includegraphics[width = 1.6in]{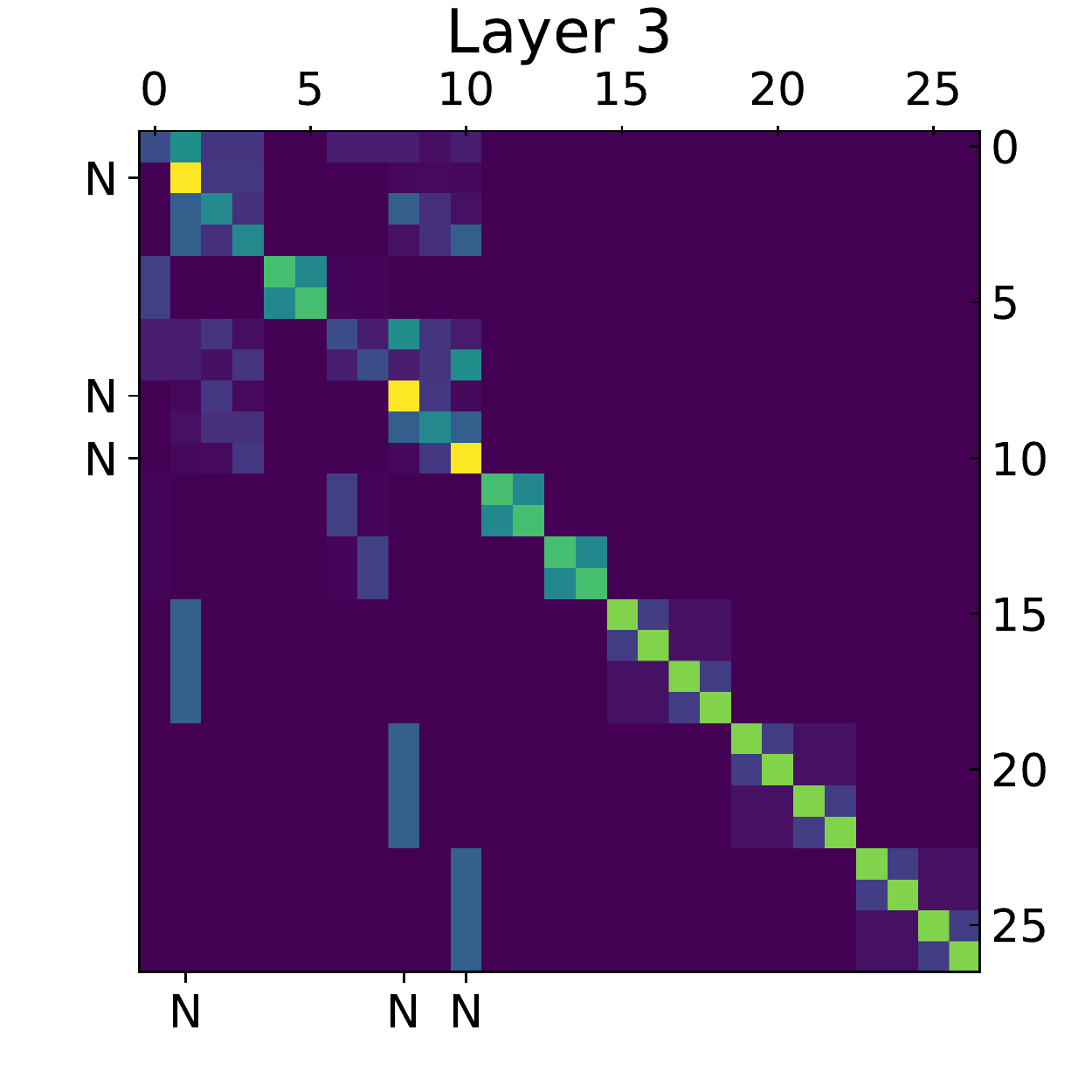}}\\
\end{tabular}
\caption{Attention scores averaged by heads for each layer of our trained model for the compound in Figure~\ref{fig:tmel}. \textit{Left}: diffusion kernel for~\ref{fig:tmel}.  \textit{Right}: node $1$, $8$, and $10$ (N) are salient.}
\label{fig:attentions_tmel}
\end{figure}

\end{document}